\documentclass[10pt,twocolumn,letterpaper]{article}

\usepackage{cvpr}              

\usepackage{graphicx}
\usepackage{amsmath}
\usepackage{amssymb}
\usepackage{booktabs}
\usepackage{enumitem, array}
\usepackage{booktabs, multirow, array, makecell, caption}
\usepackage{hhline}
\usepackage{subcaption}

\usepackage{amsmath,amsfonts,bm}

\def\eqref#1{equation~\ref{#1}}

\def\1{\bm{1}}

\DeclareMathAlphabet{\mathsfit}{\encodingdefault}{\sfdefault}{m}{sl}
\SetMathAlphabet{\mathsfit}{bold}{\encodingdefault}{\sfdefault}{bx}{n}

\usepackage{algorithm}
\usepackage{setspace}
\usepackage{algpseudocode}
\usepackage{cuted}
\usepackage{bbm}
\usepackage{float}

\usepackage[pagebackref,breaklinks,colorlinks]{hyperref}
\usepackage[accsupp]{axessibility}  

\usepackage[capitalize]{cleveref}
\crefname{section}{Sec.}{Secs.}
\Crefname{section}{Section}{Sections}
\Crefname{table}{Table}{Tables}
\crefname{table}{Tab.}{Tabs.}

\def \xplus {x^{\operatorname{aug}}}
\def \xminus {x^{\operatorname{diff}}}
\def \Sx {\mathcal{S}_x}
\def \Dx {\mathcal{D}_x}

\def \X {\mathcal{X}}

\def\cvprPaperID{11917} 
\def\confName{CVPR}
\def\confYear{2023}

\begin{document}

\title{Multiple Instance Learning via Iterative Self-Paced Supervised Contrastive Learning}

\newcommand*\samethanks[1][\value{footnote}]{\footnotemark[#1]}

\author{Kangning Liu{\thanks{Equal Contribution}}$^{*1}$, Weicheng Zhu{\samethanks}$^{*1}$, Yiqiu Shen$^{1}$,
  {Sheng Liu$^{1}$,  Narges Razavian$^{2}$} \\
  {Krzysztof J. Geras$^{\dagger 2,1}$, Carlos Fernandez-Granda{\thanks{Joint Last Author}}$^{\dagger 1,3}$} \\
  $^1$NYU Center for Data Science $^2$NYU Grossman School of Medicine \\
  $^3$ Courant Institute of Mathematical Sciences
  }
  
\maketitle
\begin{abstract} 

Learning representations for individual instances when only bag-level labels are available is a fundamental challenge in multiple instance learning (MIL). Recent works have shown promising results using contrastive self-supervised learning (CSSL), which learns to push apart representations corresponding to two different randomly-selected instances. Unfortunately, in real-world applications such as medical image classification, there is often class imbalance, so
randomly-selected instances mostly belong to the same majority class, which precludes CSSL from learning inter-class differences. To address this issue, we propose a novel framework, Iterative Self-paced Supervised Contrastive Learning for MIL Representations (\textit{ItS2CLR}), which improves the learned representation by exploiting instance-level pseudo labels derived from the bag-level labels.
The framework employs a novel self-paced sampling strategy to ensure the accuracy of pseudo labels. We evaluate \textit{ItS2CLR} on three medical datasets, showing that it improves the quality of instance-level pseudo labels and representations, and outperforms existing MIL methods in terms of both bag and instance level accuracy.  \footnote{Code is available at \url{https://github.com/Kangningthu/ItS2CLR}}

\end{abstract}

\section{Introduction}
\label{sec:newintro}
The goal of multiple instance learning (MIL) is to perform classification on data that is arranged in \emph{bags} of instances. Each instance is either positive or negative, but these instance-level labels are not available during training; only bag-level labels are available. A bag is labeled as positive if \emph{any} of the instances in it are positive, and negative otherwise. An important application of MIL is cancer diagnosis from histopathology slides. Each slide is divided into hundreds or thousands of tiles but typically only slide-level labels are available~\cite{courtiol2018classification,campanella2019clinical,li2021dual,transformermil,zhang2022dtfd,lu2021data}. 

\begin{figure*}
    \centering
    \begin{minipage}{0.51\linewidth}
    \includegraphics[width=1\linewidth]{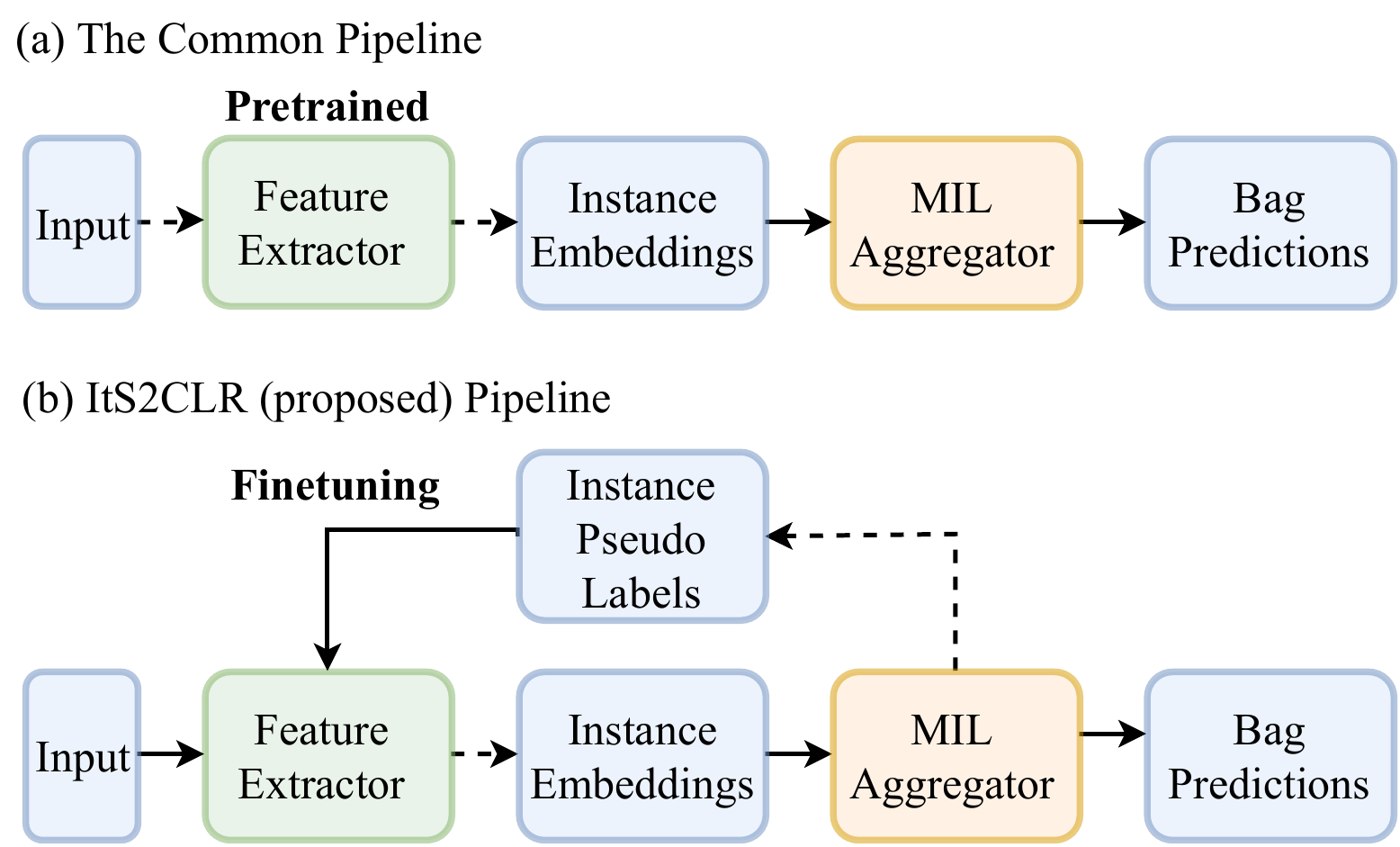}
    \end{minipage}
    \hspace{1mm}
    \begin{minipage}{0.46\linewidth}
    \includegraphics[width=1\linewidth, trim={0 0 60 0},clip]{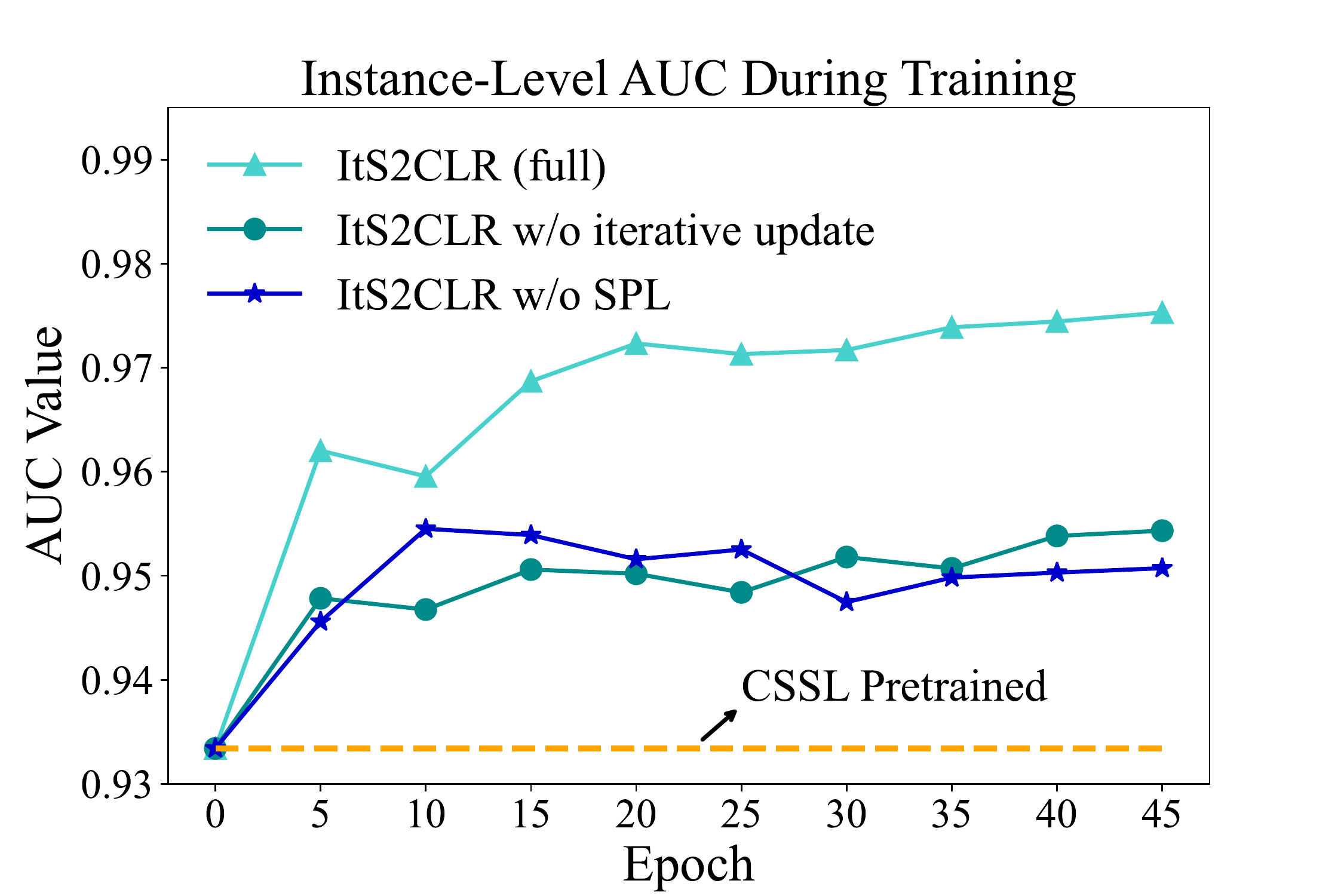}
    \end{minipage}
    \caption{\textbf{Left:} (a) Commonly used deep MIL models first pretrain a feature extractor and then train an aggregator that maps the representations to a bag-level prediction. (b) Our proposed framework, ItS2CLR, uses instance-level pseudo labels obtained from the aggregator to finetune the feature extractor. ItS2CLR updates the features iteratively based on the pseudo label of a subset of instances selected according to a self-paced learning (SPL) strategy. \textbf{Right:} The dashed line is the instance-level AUC of the MIL model trained on instance feature extracted by the CSSL pretrained feature extractor. On a benchmark dataset (Camelyon16~\cite{bejnordi2017diagnostic}), the iterative finetuning process gradually improves the instance-level AUC during training, which results in more accurate pseudo labels. Both the iterative updates and SPL are important to achieve this.
    }
    \label{fig:training_process}
\end{figure*}

Histopathology slides are typically very large, in the order of gigapixels 
(the resolution of a typical slide can be as high as $10^5 \times 10^5$), so end-to-end training of deep neural networks is typically infeasible due to memory limitations of GPU hardware.
Consequently, state-of-the-art approaches~\cite{campanella2019clinical,li2021dual,zhang2022dtfd,lu2021data, shao2021transmil} utilize a two-stage learning pipeline
: (1) a feature-extraction stage where each instance is mapped to a representation which summarizes its content, and (2) an aggregation stage where the representations extracted from all instances in a bag are combined to produce a bag-level prediction (Figure~\ref{fig:training_process}). Notably, our results indicate that even in the rare settings where end-to-end training is possible, this pipeline still tends to be superior (see Section~\ref{subsec:e2e}).

\begin{figure*}[t]
    \centering
    \includegraphics[width=1\textwidth]{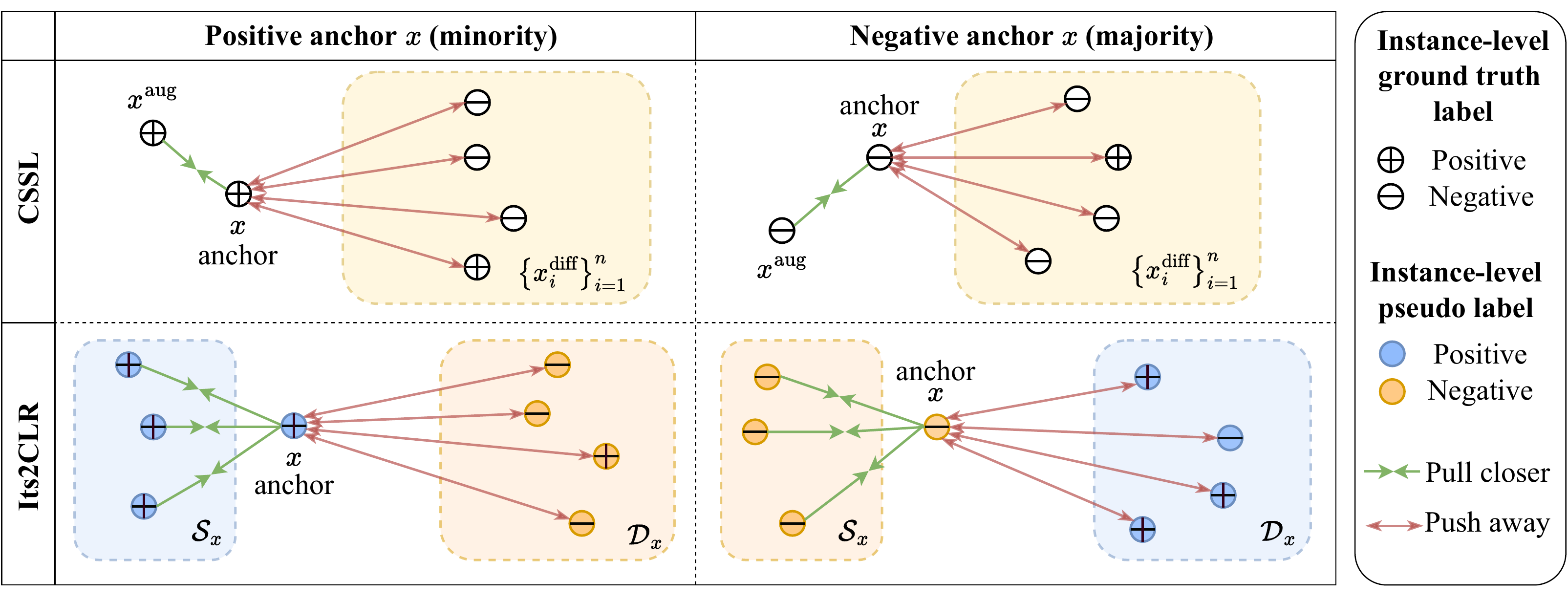}
    \caption{\textbf{Top}: In contrastive self-supervised learning (CSSL), the representation of an instance $x$ is pulled closer to its random augmentation $\xplus$ and pushed away from the representations of other randomly selected instances $\{\xminus_i\}_{i=1}^n$. In many MIL datasets relevant to medical diagnosis, most instances are negative, so CSSL mostly pushes apart representations of negative instances (right). \textbf{Bottom}: Our proposed framework ItS2CLR applies the supervised contrastive learning approach described in Section~\ref{sec:supervised_contrastive}. Instance-level pseudo labels are used to build a set of positive pairs $\Sx$ and a set of negative pairs $\Dx$ corresponding to $x$. The representation of an instance $x$ is pulled closer to those in $\Sx$ and pushed away from those in $\mathcal{D}_x$. The set of pseudo-labels is built iteratively following the self-paced sampling strategy in Section~\ref{sec:self-paced}.}
    \label{fig:sampling_stra}

\end{figure*}

In this work, we focus on a fundamental challenge in MIL: how to train the feature extractor. Currently, there are three main strategies to perform feature-extraction, which have significant shortcomings. (1) Pretraining on a large natural image dataset such as ImageNet~\cite{shao2021transmil,lu2021data} is problematic for medical applications because features learned from natural images may generalize poorly to other domains~\cite{lu2020semi}. (2) Supervised training using bag-level labels as instance-level labels is effective if positive bags contain mostly positive instances~\cite{lerousseau2020weakly,xu2019camel,chikontwe2020multiple}, but in many medical datasets this is not the case~\cite{bejnordi2017diagnostic,li2021dual}. 
(3) Contrastive self-supervised learning (CSSL) outperforms prior methods~\cite{li2021dual,ciga2022self}, but is not as effective in settings with heavy class imbalance, which are of crucial importance in medicine. 
 CSSL operates by pushing apart the representations of different randomly selected instances. When positive bags contain mostly negative instances, CSSL training ends up pushing apart negative instances from each other, which precludes it from learning features that distinguish positive samples from the negative ones (Figure~\ref{fig:sampling_stra}). We discuss this finding in Section~\ref{sec:motivation}.

Our goal is to address the shortcomings of current feature-extraction methods. We build upon several key insights. First, it is possible to extract instance-level pseudo labels from trained MIL models, which are more accurate than  {assigning the bag-level labels to all instances within a positive bag}. Second, we can use the pseudo labels to finetune the feature extractor, improving the instance-level representations. Third, these improved representations result in improved bag-level classification and more accurate instance-level pseudo labels. 
These observations are utilized in our proposed framework, Iterative Self-Paced Supervised Contrastive Learning for MIL Representation (ItS2CLR), as illustrated in Figure~\ref{fig:training_process}. After initializing the features with CSSL, we iteratively improve them via supervised contrastive learning~\cite{khosla2020supervised} using pseudo labels inferred by the aggregator. This feature refinement utilizes pseudo labels sampled according to a novel self-paced strategy, which ensures that they are sufficiently accurate  
(see Section~\ref{sec:self-paced}). In summary, our contributions are the following:

\begin{enumerate}[leftmargin=*]
    \item We propose ItS2CLR -- a novel MIL framework  where instance features are iteratively improved using pseudo labels extracted from the MIL aggregator. 
    The framework combines supervised contrastive learning with a self-paced sampling scheme to ensure that pseudo labels are accurate. 
    \item We demonstrate that the proposed approach outperforms existing MIL methods in terms of bag- and instance-level accuracy on three real-world medical datasets relevant to cancer diagnosis: two histopathology datasets and a breast ultrasound dataset. It also outperforms alternative finetuning methods, such as instance-level cross-entropy minimization and end-to-end training.
    \item In a series of controlled experiments, we show that ItS2CLR is effective when applied to different feature-extraction architectures and when combined with different aggregators. 
\end{enumerate}

\section{CSSL May Not Learn Discriminative Representations In MIL} 
\label{sec:motivation}
Recent MIL approaches use contrastive self-supervised learning (CSSL) to train the feature extractor~\cite{li2021dual,dehaene2020self, Rymarczyk2021ProtoMILMI}. In this section, we show that CSSL (e.g. SimCLR~\cite{chen2020simple}, MoCo~\cite{he2020momentum}) has a crucial limitation in realistic MIL settings, which precludes it from learning discriminative features. CSSL aims to learn a representation space where samples from the same class are close to each other, and samples from different classes are far from each other, without access to class labels. This is achieved by minimizing the InfoNCE loss~\cite{oord2018representation}.

\begin{equation}
\label{eq:ssl}
\footnotesize
\mathcal{L}_{\text{CSSL}} = \underset{\underset{\{\xminus_i\}_{i=1}^n}{x, \xplus}}{\mathbb{E}} \left [ -\log \frac{sim(x, \xplus)}{sim(x, \xplus) +\sum_{i=1}^n sim(x, \xminus_i)}\right].
\end{equation}
The similarity score 
\smash{$sim\left(\cdot, \cdot \right) : \mathbb{R}^m \times \mathbb{R}^m \rightarrow \mathbb{R}$} is defined as \smash{$ sim\left(x, x' \right) = \exp \left(f_\psi(x) \cdot f_\psi(x') / \tau \right)$} for any \smash{$x, x' \in \mathbb{R}^m$}, 
where  $f_\psi = \psi \circ f$, in which $f:\mathbb{R}^m\rightarrow \mathbb{R}^d$ is the feature extractor mapping the input data to a representation, $\psi: \mathbb{R}^d \rightarrow \mathbb{R}^{d'}$ is a projection head with a feed-forward network and $\ell_2$ normalization, and $\tau$ is a temperature hyperparameter. The expectation is taken over samples $x \in \mathbb{R}^m$ drawn uniformly from the training set.  Minimizing the loss brings the representation of an instance $x$ closer to the representation of its random augmentation, $\xplus$, and pushes the representation of $x$ away from the representations of $n$ other examples $\{\xminus_i\}_{i=1}^n$ in the training set. 

A key assumption in CSSL is that $x$ belongs to a different class than most of the randomly-sampled examples \smash{$\xminus_1$}, \ldots, \smash{$\xminus_n$}. This usually holds in standard classification datasets with many classes such as ImageNet~\cite{deng2009imagenet}, but \emph{not in MIL tasks relevant to medical diagnosis}, where a majority of instances are negative (e.g. 95\% in Camelyon16). Hence, most terms in the sum \smash{$\sum_{i=1}^n \exp \left(f_\psi(x) \cdot f_\psi(\xminus_i)/ \tau\right)$} in the loss in Equation~\ref{eq:ssl} correspond to pairs of examples $(x,\xminus_i)$ both belonging to the negative class. Therefore, minimizing the loss mostly pushes apart the representations of negative instances, as illustrated in the top panel of Figure~\ref{fig:sampling_stra}. This is an example of \textit{class collision}~\cite{arora2019theoretical, chuang2020debiased}, a general problem in CSSL, which has been shown to impair performance on downstream tasks~\cite{ash2021investigating,zheng2021weakly}.

Class collision makes CSSL learn representations that are not discriminative between classes. In order to study this phenomenon, we report the average inter-class distances and intra-class deviations for representations learned by CSSL on Camelyon16 in Table~\ref{tab:calsscollision}. The inter-class distance reflects how far the instance representations from different classes are apart; the intra-class distance reflects the variation of instance representations within each class. As predicted, the intra-class deviation corresponding to the representations of negative instances learned by CSSL is large. 
Representations learned by ItS2CLR have larger inter-class distance (more separated classes) and smaller intra-class deviation (less variance among instances belonging to the same class) than those learned by CSSL.
This suggests that the features learned by ItS2CLR are more discriminative, which is confirmed by the results in Section~\ref{sec:experiments}.

\begin{table*}[t]
    \centering
    \caption{Quantitative analysis of instance-level features learned from Camelyon16~\cite{bejnordi2017diagnostic}. The inter-class distance is the $\ell_2$-distance between the representation mean of the positive instances and that of negative instances. The intra-class deviation is the square root of the spectral norm of the covariance matrix of features corresponding to each class.
    The spectral norm is the largest eigenvalue of the covariance matrix and is therefore equal to the variance in the direction of the greatest variance.
    Due to class collision among negative instances in CSSL (see Section~\ref{sec:motivation}), the intra-class deviation of the corresponding features is very large. In contrast, the features learned by
    the proposed framework ItS2CLR has smaller intra-class deviation among both negative and positive instances, and a larger inter-class distance.}
    \renewcommand{\arraystretch}{1.15}
    \resizebox{0.8\linewidth}{!}{
    \begin{tabular}{l|c|cc|c|cc}
    \toprule
     &\multicolumn{3}{c|}{Training set}  & \multicolumn{3}{c}{Test set} \\
        \cline{2-7}
    & \multirow{2}{*}{Inter-class distance}  & \multicolumn{2}{c|}{Intra-class deviation} & \multirow{2}{*}{Inter-class distance}  & \multicolumn{2}{c}{Intra-class deviation} \\
    \cline{3-4} \cline{6-7}
    & &    \textit{pos}   & \textit{neg} & & \textit{pos}   & \textit{neg}\\ 
    \hline
{CSSL (SimCLR)}  & 1.835   &      1.299 &      1.453  & 2.109   &      1.416   &      1.484\\
{ItS2CLR (proposed)} & \textbf{2.376}   &      \textbf{1.176}  &      \textbf{0.805}  &  \textbf{2.432}    &     \textbf{1.215}    &    \textbf{ 0.847}  \\
    \bottomrule
    \end{tabular}
    }
    \label{tab:calsscollision}
\end{table*}

Note that using bag-level labels does not solve the problem of class collision.
When $x$ is negative, even if we select $\{\xminus_i\}_{i=1}^n$ from the positive bags in \eqref{eq:ssl}, most of the selected instances will still be negative. Overcoming the class-collision problem requires explicitly detecting positive instances. This motivates our proposed framework, described in the following section.

\section{MIL via Iterative Self-paced Supervised Contrastive learning}
\label{sec:method}

\textbf{It}erative \textbf{S}elf-paced \textbf{S}upervised \textbf{C}ontrastive \textbf{L}earning for MIL \textbf{R}epresentations (ItS2CLR) addresses the limitation of contrastive self-supervised learning (CSSL) described in Section~\ref{sec:motivation}. ItS2CLR relies on latent variables indicating whether each instance is positive or negative, which we call \emph{instance-level pseudo labels}. To estimate pseudo labels, we use instance-level probabilities obtained from the MIL aggregator (specifically the aggregator from DS-MIL~\cite{li2021dual} but our framework is compatible with any aggregator that generates instance-level probabilities, such as the ones described in Appendix~\ref{app:mil_agg}). The pseudo labels are obtained by binarizing the probabilities according to a threshold $\eta \in (0,1)$, which is a hyperparameter.

ItS2CLR uses the pseudo labels to finetune the feature extractor (initialized using CSSL). {In the spirit of iterative self-training techniques~\cite{zhong2019graph, wei2020theoretical, liu2021adaptive}}, we alternate between refining the feature extractor, re-computing the pseudo labels, and training the aggregator, as described in Algorithm~\ref{algo:main}. 
A key challenge is that the pseudo labels are not completely accurate, especially at the beginning of the training process. To address the impact of incorrect pseudo labels, we apply a contrastive loss to finetune the feature extractor (see Section~\ref{sec:supervised_contrastive}), where the contrastive pairs are selected according to a novel self-paced learning scheme (see Section~\ref{sec:self-paced}). 
The right panel of Figure~\ref{fig:training_process} shows that our approach iteratively improves the pseudo labels on the Camelyon16 dataset~\cite{bejnordi2017diagnostic}.
 {This finetuning only requires a modest increment in computational time (see Section~\ref{app:computationtime}}).

\subsection{Supervised contrastive learning with pseudo labels}
\label{sec:supervised_contrastive}
To address the class collision problem described in Section~\ref{sec:motivation}, we leverage \emph{supervised} contrastive learning~\cite{pantazis2021focus, dwibedi2021little, khosla2020supervised} combined with the pseudo labels estimated by the aggregator. The goal is to learn discriminative representations by pulling together the representations corresponding to instances in the same class, and pushing apart those belong to instances of different classes. For each \textit{anchor} instance $x$ selected for contrastive learning, we collect a set $\Sx$ believed to have the \emph{same label} as $x$, and a set $\Dx$ believed to have a \emph{different label} to $x$. These sets are depicted in the bottom panel of Figure~\ref{fig:sampling_stra}.
The supervised contrastive loss corresponding to $x$ is defined as:

\begin{equation}
\footnotesize
\mathcal{L}_{\text {sup }}(x) = \frac{1}{|\Sx|} \sum_{x_s \in \Sx} - \log \frac{sim\left(x, x_s \right)}{\sum\limits_{x_s \in \Sx} sim\left(x, x_s \right) + \sum\limits_{x_d \in \Dx} sim\left(x, x_d \right)}.
\label{eq:sup}
\end{equation}

In Section~\ref{sec:self-paced}, we explain how to select $x$, $\Sx$ and $\Dx$ to ensure that the  selected samples have high-quality pseudo labels.

A tempting alternative to supervised constrastive learning is to train the feature extractor on pseudo labels using standard cross-entropy loss. However, in Section~\ref{subsec:e2e} we show that this leads to substantially worse performance in the downstream MIL classification task due to memorization of incorrect pseudo labels.

\let\Algorithm\algorithm
\renewcommand\algorithm[1][]{\Algorithm[#1]\setstretch{1.1}}
\algrenewcommand\algorithmiccomment[1]{ {\itshape \textcolor{blue}{ \# #1}}}

\begin{algorithm}[H]
\caption{ Iterative Self-Paced Supervised Contrastive Learning (It2SCLR)}\label{alg:cap}
\begin{algorithmic}[1]
\footnotesize

\Require Feature extractor $f$, projection head $\psi$;
\Require MIL aggregator $g_{\phi}$, where $\phi$ is an instance classifier;
\Require Bags $ \{X_b\}_{b=1}^{B}$, bag-level labels $\{Y_b\}_{b=1}^{B}$;

\State${f}^{(0)} \gets f_{\text{SSL}}$  \Comment{Initialize $f$ with SSL-pretrained weights}
\For{$t =  0$ to T}
\State \Comment{Extract instance representation}
\State  ${h_k^b} \gets {f^{(t)}(x_k^b), \forall x_k^b \in X_b, \forall b}$  
\State \Comment{Group instance embedding into bags}
\State  ${H_b \gets {\{h_k^b\}_{k=1}^{K_b}}, \forall b}$ 
\State \Comment{Train the MIL aggregator}
\State $g_{\phi}^{(t)} \gets$ Train with $\{H_b\}_{b=1}^{B}$ and $\{Y_b\}_{b=1}^{B}$ 
\State $\text{AUC}^{(t)}_{val} \gets$ bag-level AUC on the validation set
\State \Comment{If the bag prediction improves on the validation set} 
\If{${\text{AUC}^{(t)}_{val} \ge \max_{t' \le t} \left\{ \text{AUC}^{(t')}_{val} \right\}}$} 
\State \Comment{Update instance pseudo labels}
\State $\hat{y}_k^b \gets \mathbbm{1}_{\{\phi^{(t)}(h_k^b) > \eta \}}, \forall x_k^b \in X_b, \forall b$ 
\EndIf
\State \Comment{Optimize feature extractor via Eq.(\ref{eq:sup})}
\State \smash{$f_\psi^{(t+1)} \gets \text{argmin}_{f_\psi} \mathcal{L}_{\text{sup}}(f_\psi^{(t)})$} 
\EndFor
\end{algorithmic}
\label{algo:main}

\end{algorithm}

\subsection{Sampling via self-paced learning}
\label{sec:self-paced}

A key challenge in It2SCLR is to improve the accuracy of instance-level pseudo labels without ground-truth labels. 
This is achieved by finetuning the feature extractor on a carefully-selected subset of instances. We select the anchor instance $x$ and the corresponding sets $\Sx$ and $\Dx$ (defined in Section~\ref{sec:supervised_contrastive} and Figure~\ref{fig:sampling_stra}) building upon two key insights: (1) The negative bags only contain negative instances. (2) The probabilities used to build the pseudo labels are indicative of their quality; instances with higher predicted probabilities usually have more accurate pseudo labels~\cite{zou2018unsupervised,liu2020early,liu2021adaptive}. 

Let \smash{$\X^-_{\text{neg}}$} denote all instances within the negative bags. By definition of MIL, we can safely assume that all instances in \smash{$\X^-_{\text{neg}}$} are negative. In contrast, positive bags contain both positive and negative instances. Let \smash{$\X_{\text{pos}}^+$} and \smash{$\X_{\text{pos}}^-$} denote the sets of instances in positive bags with positive and negative pseudo labels respectively. During an initial warm-up lasting \smash{$T_{\text{warm-up}}$} epochs, we sample anchor instances $x$ only from \smash{$\X^-_{\text{neg}}$} to ensure that they are indeed all negative. For each such instance, \smash{$\Sx$} is built by sampling instances from \smash{$\X^-_{\text{neg}}$}, and \smash{$\Dx$} is built by sampling from \smash{$\X_{\text{pos}}^+$}.

After the warm-up phase, we start sampling anchor instances from \smash{$\X_{\text{pos}}^+$} and \smash{$\X_{\text{pos}}^-$}. To ensure that these instances have accurate pseudo labels, we only consider the  
top-$r\%$ instances with the highest confidence in each of these sets (i.e. the highest probabilities in \smash{$\X_{\text{pos}}^+$} and lowest probabilities in \smash{$\X_{\text{pos}}^-$}), which we call \smash{$\X_{\text{pos}}^+(r)$} and \smash{$\X_{\text{pos}}^-(r)$} respectively, as illustrated by Appendix Figure~\ref{fig:wholeset}. The ratio of positive-to-negative anchors is a fixed hyperparameter $p_{+}$.
For each anchor $x$, the same-label set \smash{$\Sx$} is sampled from \smash{$\X_{\text{pos}}^+(r)$} if $x$ is positive and from \smash{$\X^-_{\text{neg}} \cup \X_{\text{pos}}^-(r)$} if $x$ is negative. The different-label set $\Dx$ is sampled from 
\smash{$\X^-_{\text{neg}} \cup \X_{\text{pos}}^-(r)$} if $x$ is positive, and from \smash{$\X_{\text{pos}}^+(r)$} if $x$ is negative.

\begin{table*}[t]
    \centering
    \caption{Bag-level AUC of ItS2CLR and a two-stage baseline using a SimCLR feature extractor and a MIL aggregator.
    ItS2CLR outperforms the baseline on all three datasets.}
    \renewcommand{\arraystretch}{1.05}
        \resizebox{0.8\linewidth}{!}{
    \begin{tabular}{l|>{\centering\arraybackslash}m{0.15\linewidth}|>{\centering\arraybackslash}m{0.15\linewidth}|cccc}
    \toprule
    AUC $(\times 10^{-2})$ & \multirow{2}{*}{Camelyon16} & \multirow{2}{*}{Breast Ultrasound} & \multicolumn{4}{c}{TCGA-LUAD mutation}\\
    & & &  EGFR  & KRAS  & STK11 & TP53\\
    \hline
SimCLR + Aggregator  & 85.38 & 80.79  & 67.51  & 68.79 & 70.40 & 62.15 \\
ItS2CLR &\textbf{94.25} & \textbf{93.93} & \textbf{72.30}  & \textbf{71.06} & \textbf{75.08} & \textbf{65.61} \\
    \bottomrule
    \end{tabular}
    }
    \label{tab:mainresult}
\end{table*}

To exploit the improvement of the instance representations during training, we gradually increase $r$ to include more instances from positive bags, 
which can be interpreted as a self-paced \textit{easy-to-hard} learning scheme~\cite{kumar2010self,jiang2014self,zou2018domain}. Let $t$ and $T$ denote the current epoch, and the total number of epochs respectively. 
For  \smash{$T_{\text{warmup}} < t \leq T$}. we set:
\begin{equation}
    r :=r_{0} + \alpha_{r}\left(t-T_{\text{warm-up}}\right),
\end{equation}
where  \smash{$ \alpha_{r}={(r_{T}-r_{0})}/{(T-T_{\text{warm-up}})}$}, $r_{0}$ and $r_{T}$ are hyperparameters. Details on tuning these hyperparameters are provided in Appendix~$\ref{app:hyperparameter}$. As demonstrated in the right panel of Figure~\ref{fig:training_process} (see also Appendix~\ref{app:learning_curves}), this scheme indeed results in an improvement of the pseudo labels (and hence of the underlying representations).

\section{Experiments}
\label{sec:experiments}

We evaluate {ItS2CLR} on three MIL datasets described in Section~\ref{subsec:exp_setip}. 
In Section~\ref{subsec:aggregator} we show that ItS2CLR consistently outperforms approaches that use CSSL feature-extraction by a substantial margin on all three datasets for different choices of aggregators. 
In Section~\ref{subsec:e2e} we show that ItS2CLR outperforms alternative finetuning approaches based on cross-entropy loss minimization and end-to-end training across a wide range of settings where the prevalence of positive instances and bag size vary.  {In Section~\ref{sec:otherinitializations}, we show that ItS2CLR is able to improve features obtained from a variety of pretraining schemes and network architectures.}

\begin{table*}[t]
    \centering
    \caption{Bag-level AUC on Camelyon16 for ItS2CLR and different baselines for five aggregators. We retrain each aggregator 5 times to report the mean and standard deviation (reported as a subscript). All feature extractors are initialized using SimCLR. Ground-truth and cross-entropy (CE) finetuning use ground-truth instance-level labels and pseudo labels to optimize the feature extractor respectively. We also include versions of ItS2CLR without iterative updates (w/o iter.), self-paced learning (w/o SPL) and both (w/o both), and a version of CE finetuning without iterative updates (w/o iter). { See Appendix~\ref{Detailsablationmodel} for a detailed description of the ablated models.}
    }
    \resizebox{\linewidth}{!}{
    \renewcommand{\arraystretch}{1.05}
    \begin{tabular}{lcc|cc|cccc}
    \toprule
    AUC $(\times 10^{-2})$& \multirow{2}{1.5cm}{\centering{SimCLR (CSSL)}} & \multirow{2}{2cm}{\centering{Ground-truth finetuning*}} & \multicolumn{2}{c|}{CE finetuning} & \multicolumn{4}{c}{ItS2CLR} \\
    &&& \textit{w/o iter.} & \textit{iter.} & \textit{w/o both} & \textit{w/o iter.}  & \textit{w/o SPL}   & \textit{Full}  \\
    \midrule
    Max pooling      & $86.69_{1.09}$ & $98.25_{0.01}$ & $85.48_{0.24}$ & $88.05_{0.77}$ & $85.38_{0.31}$ & $91.96_{0.31}$ & $90.85_{0.76}$ & $\textbf{94.69}_{0.07}$\\
    Top-k pooling~\cite{shen2021interpretable} & $85.39_{1.20}$ & $98.39_{0.05}$ & $85.96_{0.45}$ & $87.26_{0.42}$ & $85.46_{0.21}$ & $91.73_{0.42}$ & $91.69_{0.28}$ & $\textbf{95.07}_{0.09}$\\
    Attention-MIL~\cite{ilse2018attention} & $79.49_{3.20}$ & $99.06_{0.02}$ & $88.50_{0.54}$ & $90.46_{0.64}$ & $85.21_{0.74}$ & $93.13_{0.22}$ & $86.20_{3.25}$ & $\textbf{94.45}_{0.05}$\\
    DS-MIL~\cite{li2021dual}  & $85.38_{1.32}$ & $98.65_{0.08}$ & $87.01_{0.82}$ & $90.38_{0.67}$ & $85.08_{0.38}$ & $91.69_{0.54}$ & $88.29_{0.99}$ & $\textbf{94.25}_{0.07}$  \\
    Transformer~\cite{transformermil} & $87.25_{0.59}$ & $98.85_{0.25}$ & $89.02_{0.54}$ & $92.13_{1.07}$ & $87.13_{0.71}$ & $93.52_{0.49}$ & $92.12_{0.68}$ & $\textbf{95.74}_{0.27}$  \\
\bottomrule
\end{tabular}
}
    \label{tab:aggregator_results}
\end{table*}

\begin{table*}[ht]
    \centering
    \caption{Comparison of instance-level performance between the models in Table~\ref{tab:aggregator_results}. All models use the DS-MIL aggregator. ItS2CLR achieves the best localization performance. Dice score is computed from a post-processed probability estimate described in Appendix~\ref{app:ins_eval}, which also includes further details and results for other aggregators.}
    \resizebox{0.9\linewidth}{!}{
    \renewcommand{\arraystretch}{1}
    \begin{tabular}{lcc|cc|cccc}
    \toprule
     $(\times 10^{-2})$& SimCLR & \multirow{2}{2cm}{\centering{Ground-truth finetuning}} & \multicolumn{2}{c|}{CE finetuning} & \multicolumn{4}{c}{ItS2CLR} \\
    & (CSSL) && \textit{w/o iter.} & \textit{iter.} & \textit{w/o both} & \textit{w/o iter.}  & \textit{w/o SPL}   & \textit{Full}  \\
    \midrule
    AUC &94.01 & 97.94 & 95.69 & 96.06 & 95.13 & 95.90 & 96.12 & \textbf{96.72}\\
    F1-score     &84.49 & 88.01 & 86.94 & 86.93 & 86.74 & 87.45 & \textbf{87.95} & 87.47\\
    AUPRC  & 86.57 & 86.13 & 89.26 & 89.39 & 88.30 & 89.51 & 90.00 & \textbf{91.12} \\
    Dice (*) & 31.79 & 62.17 & 49.11 & 49.41 & 43.74 & 51.70 & 53.03 & \textbf{57.86} \\
    IoU & {39.53} & {50.24} & {44.98} & {44.88} & {41.37} & {44.56} & {45.41} & {\textbf{48.27}} \\
\bottomrule
\end{tabular}
}
\label{tab:instance_test}
\end{table*}

\begin{figure*}[t]
    \centering
    \includegraphics[width=\textwidth]{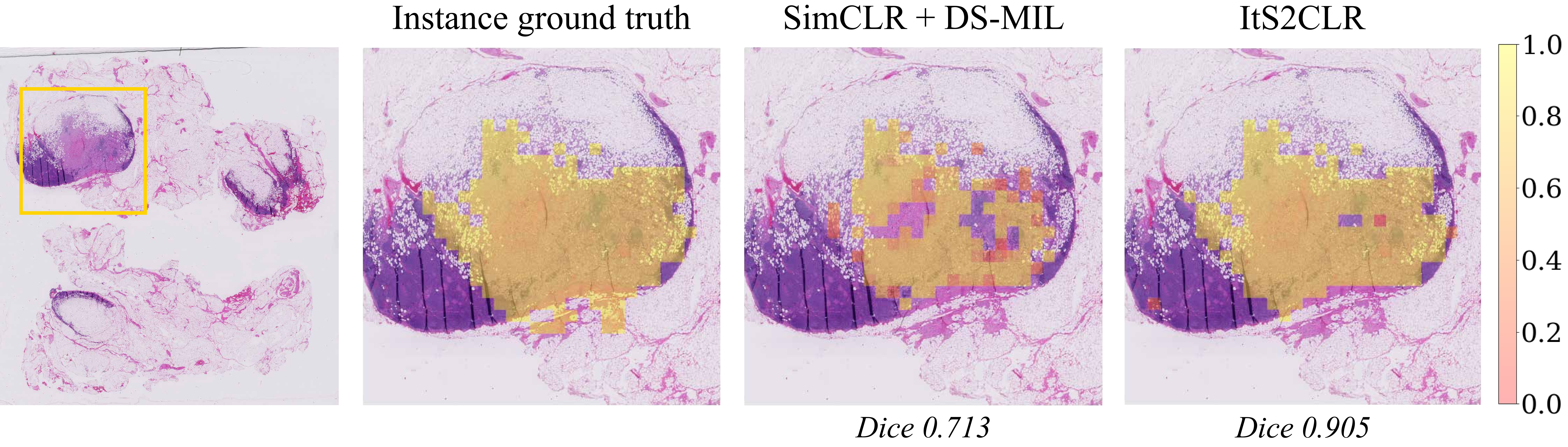}
    \vspace{-4mm}
    \caption{Tumor localization in a histopathology slide from the Camelyon16 test set. Instance-level predictions are generated by the instance-level classifier of the DS-MIL aggregator based on different instance representations. Yellow indicates higher probability of being cancerous. Transparent tiles are with probabilities less than 0.2.
    Appendix~\ref{app:localization_map} shows additional examples.}
    \vspace{-5mm}
    \label{fig:localization}
\end{figure*}

\begin{table*}[t]
\caption{Comparison to models trained end-to-end, initialized with the same pretrained weights as ItS2CLR, and use the same aggregator. ItS2CLR achieves better instance- and bag-level performance.}
\centering
\resizebox{\linewidth}{!}{
\centering
\begin{tabular}{>{\centering\arraybackslash}m{0.13\linewidth} | >{\centering\arraybackslash}m{0.1\linewidth} >{\centering\arraybackslash}m{0.1\linewidth} >{\centering\arraybackslash}m{0.1\linewidth}|
>{\centering\arraybackslash}m{0.1\linewidth} >{\centering\arraybackslash}m{0.1\linewidth} >{\centering\arraybackslash}m{0.1\linewidth}
>{\centering\arraybackslash}m{0.1\linewidth}}
\toprule
&  \multicolumn{3}{c|}{Camelyon16 (downsampled synthetic)} & \multicolumn{4}{c}{Breast Ultrasound} \\
\midrule[0.7pt]
 & {Bag AUC} & {Instance AUC} & {Instance F-score } &{Bag AUC} & {Bag AUPRC} & {Instance AUC} & {Instance AUPRC}\\
\midrule[0.7pt]
 End-to-end & 66.71 & 78.32 & 55.71 & 91.26 & 58.73 & 82.11 & 31.31 \\
ItS2CLR  & \textbf{88.65} & \textbf{95.58} &\textbf{87.01} & \textbf{93.93} & \textbf{70.30} &\textbf{88.63} & \textbf{43.71}\\
 \bottomrule
\end{tabular}
}
\vspace{-3mm}
\label{tab:e2e_comp}
\end{table*}

\begin{table}[t]
    \centering
    \caption{Bag-level AUC on Camelyon16 across different witness rates (WR): the fraction of positive instances in positive bags. All methods use DS-MIL aggregator for a fair comparison.
    When the WR is low, ItS2CLR outperforms CE finetuning by a large margin. 
    As the WR increases, CE finetuning becomes more effective.}
    \renewcommand{\arraystretch}{1.05}
    \resizebox{0.47\textwidth}{!}{
    \begin{tabular}{>{\centering\arraybackslash}m{0.2\linewidth} | >{\centering\arraybackslash}m{0.12\linewidth} | >{\centering\arraybackslash}m{0.16\linewidth} >{\centering\arraybackslash}m{0.24\linewidth}
     >{\centering\arraybackslash}m{0.16\linewidth} |
    >{\centering\arraybackslash}m{0.24\linewidth}}
    \toprule Down. Instances & WR (\%) & SimCLR (CSSL)  & CE finetuning iterative & ItS2CLR & Finetuning w. instance GT \\
    \midrule 
    $5\%$ Neg. & $71.2$ & 94.52 & \textbf{98.55} & {97.58} & 99.11\\
    $10\%$ Neg. & $45.0$ & 93.70 & \textbf{97.88} &    {96.15} & 99.18\\
    $40\%$ Neg. & $23.5$ & 90.38 & 93.32 & \textbf{95.40} & 97.68 \\
    Original & $10.9$ & 85.38 & 90.38 & \textbf{94.25} & 98.65\\
    $50\%$ Pos. & $5.8$ & 82.47 & 86.96  &  \textbf{88.52} & 91.81\\
    $33\%$ Pos. & $4.1$ & 78.21 & 80.56  & \textbf{86.02} & 88.01\\
    \bottomrule
    
    \end{tabular}
    }
    \label{tab:WR_exp}
    
\end{table}

\subsection{Datasets}
\label{subsec:exp_setip}
We evaluate the proposed framework on three cancer diagnosis tasks. When training our models, we select the model with the highest bag-level performance on the validation set and report the performance on a held-out test set. More information about the datasets, experimental setup, and implementation is provided in Appendix~\ref{app:experiments}.

\textbf{Camelyon16}~\cite{bejnordi2017diagnostic} is a popular benchmark for MIL~\cite{li2021dual, shao2021transmil,zhang2022dtfd} where the goal is to detect breast-cancer metastasis in lymph node sections. It consists of 400 whole-slide histopathology images. Each whole slide image (WSI) corresponds to a bag with a binary label indicating the presence of cancer. Each WSI is divided into an average of 625 tiles at 5x magnification, which correspond to individual instances. 
The dataset also contains pixel-wise annotations indicating the presence of cancer, which can be used to derive ground-truth instance-level labels.

\textbf{TCGA-LUAD} is a dataset from The Cancer Genome Atlas (TCGA)~\cite{tcga}, a landmark cancer genomics program, where the associated task is to detect genetic mutations in cancer cells. We build models to detect four mutations - EGFR, KRAS, STK11, and TP53, which are important to determine treatment options for LUAD~\cite{Coudray197574, Fu813543}. The data contains 800 labeled tumorous frozen WSIs from lung adenocarcinoma (LUAD). Each WSI is divided into an average of 633 tiles at 10x magnification corresponding to unlabeled instances.

The \textbf{Breast Ultrasound Dataset} contains 28,914 B-mode breast ultrasound exams~\cite{shen2021artificial}. The associated task is to detect breast cancer. Each exam contains between 4-70 images (18.8 images per exam on average) corresponding to individual instances, but only a bag-level label indicating the presence of cancer is available per exam. Additionally, a subset of images is annotated, which makes it possible to also evaluate instance-level performance. This dataset is imbalanced at the bag level:  {only 5,593 of 28,914 exams contain  cancer.}

\subsection{Comparison with contrastive self-supervised learning} 
\label{subsec:aggregator}
In this section, we compare the performance of ItS2CLR to a baseline that performs feature-extraction via the CSSL method SimCLR~\cite{chen2020simple}. This approach has achieved state-of-the-art performance on multiple WSI datasets~\cite{li2021dual}. 
To ensure a fair comparison, we initialize the feature extractor in ItS2CLR also using SimCLR. 
Table~\ref{tab:mainresult} shows that ItS2CLR clearly outperforms the SimCLR baseline on all three datasets. The performance improvement is particularly significant in Camelyon16 where it achieves a bag-level AUC of 0.943, outperforming the baseline by an absolute margin of 8.87\%. ItS2CLR also outperforms an improved baseline reported by Li \etal~\cite{li2021dual} with an AUC of 0.917, which uses higher resolution tiles than in our experiments (both 20x and 5x, as opposed to only 5x).

To perform a more exhaustive comparison of the features learned by SimCLR and ItS2CLR, we compare them in combination with several different popular MIL aggregators.\footnote{To be clear, the ItS2CLR features are learned using the DS-MIL aggregator, as described in Section~\ref{sec:method}, and then frozen before combining them with the different aggregators.}: max pooling, top-k pooling~\cite{shen2021interpretable}, attention-MIL pooling~\cite{ilse2018attention}, DS-MIL pooling~\cite{li2021dual}, and transformer~\cite{transformermil} (see Appendix~\ref{app:mil_agg} for a detailed description). 
Table~\ref{tab:aggregator_results} shows that the ItS2CLR features outperform the SimCLR features by a large margin for all aggregators, and are substantially more stable (the standard deviation of the AUC over multiple trials is lower).

 {We also evaluate instance-level accuracy, which can be used to interpret the bag-level prediction (for instance, by revealing tumor locations).} In Table~\ref{tab:instance_test}, we report the instance-level AUC, F1 score, and Dice score of both ItS2CLR and the SimCLR-based baseline on Camelyon16. 
ItS2CLR again exhibits stronger performance. 
Figure~\ref{fig:localization} shows an example of instance-level predictions in the form of a tumor localization map.

\subsection{Comparison with alternative approaches}
\label{subsec:e2e}
In Tables~\ref{tab:aggregator_results}, \ref{tab:instance_test} and \ref{tab:e2e_comp}, we compare ItS2CLR with the approaches described below. Table~\ref{tab:WR_exp} reports additional comparisons at different \emph{witness rates} (the fraction of positive instances in positive bags), created synthetically by modifying the ratio between negative and positive instances in Camelyon16.

\textbf{Finetuning with ground-truth instance labels} provides an upper bound on the performance that can be achieved through feature improvement. ItS2CLR does not reach this gold standard, but substantially closes the gap.

\textbf{Cross-entropy finetuning with pseudo labels}, which we refer to as {\textit{CE finetuning}}, consistently underperforms ItS2CLR when combined with different aggregators, except at high witness rates. We conjecture that this is due to the sensitivity of the cross-entropy loss to incorrect pseudo labels. We experiment with two settings: CE finetuning \textit{without} and \textit{with} iterative updating. In CE finetuning \textit{without} iterative updating, we use the same initial pseudo labels and pretrained representations as our ItS2CLR framework. Concretely, we label all the instances in negative bags as negative, and the instances in positive bags using the instance prediction obtained from the initially trained MIL aggregator. When finetuning the feature extractor, pseudo labels are kept fixed. In CE finetuning \textit{with} iterative updating, the pseudo labels are updated every few epochs. 

\textbf{Ablated versions of ItS2CLR} where we do not apply iterative updates of the pseudo labels (w/o iter.), or our self-paced learning scheme (w/o SPL) or both (w/o both) achieve substantially worse performance than the full approach. This indicates that both of these ingredients are critical in learning discriminative features.

\textbf{End-to-end training} is often computationally infeasible in medical applications. We compare ItS2CLR to end-to-end models on a downsampled version of Camelyon16 (see Appendix~\ref{app:subbag}) and on the breast ultrasound dataset. For a fair comparison, all end-to-end models use the same CSSL-pretrained weights and aggregator as used in ItS2CLR. Table~\ref{tab:e2e_comp} shows that ItS2CLR achieves better instance- and bag-level performance than end-to-end training.  In Appendix~\ref{append:e2e} we show that end-to-end models suffer from overfitting.

\begin{table}[h]
\caption{The effects of finetuning from different initial weights on Camelyon16. \textbf{(a)} The bag-level prediction performance on \textbf{test} set; \textbf{(b)} the evaluation on instance-level prediction in \textbf{training} set. }
    \begin{subtable}{\linewidth}
      \centering
        \renewcommand{\arraystretch}{1}
    \begin{tabular}{l|c|c|c|c}
    \toprule
     \multirow{2}{*}{\parbox{1.5cm}{Bag-level test AUC}} & \multirow{2}{*}{ImageNet} & \multirow{2}{*}{BYOL}   & \multirow{2}{*}{SimCLR} & 
     DINO
\\ 
 & & & & (ViT) \\
    \midrule
    Pretrained & 0.712 & 0.704  & 0.854 & 0.857
 \\
    ItS2CLR & \textbf{0.791} & \textbf{0.764}  & \textbf{0.943} & \textbf{0.936}
\\
    \bottomrule
    \end{tabular}
    \caption{}
    \label{tab:diffpretrained_bag}
    \end{subtable}
    \\
    \begin{subtable}{\linewidth}
      \centering
        \resizebox{1\textwidth}{!}{
            \begin{tabular}{l| >{\centering\arraybackslash}m{0.06\linewidth} >{\centering\arraybackslash}m{0.134\linewidth} | >{\centering\arraybackslash}m{0.06\linewidth} >{\centering\arraybackslash}m{0.134\linewidth} |>{\centering\arraybackslash}m{0.06\linewidth} >{\centering\arraybackslash}m{0.134\linewidth} |>{\centering\arraybackslash}m{0.06\linewidth} >{\centering\arraybackslash}m{0.134\linewidth}}
            \toprule
                 \multirow{2}{*}{\parbox{2cm}{Training ins. pred}} & \multicolumn{2}{c}{ImageNet}                & \multicolumn{2}{c}{BYOL}                    & \multicolumn{2}{c}{SimCLR}                  & \multicolumn{2}{c}{DINO (ViT)}                    \\
                  & \textit{AUC}         & \textit{{F-score}}     & \textit{AUC}         & \textit{F-score}     & \textit{AUC}         & \textit{F-score}     & \textit{AUC}         & \textit{F-score}     \\
            \midrule
        Initial   & 0.772                & 0.314                & 0.729                & 0.225                & 0.934                & 0.746                & 0.906                & 0.716                \\
        Finetuned & \textbf{0.835}                & \textbf{0.598}                & \textbf{0.804}                & \textbf{0.288}                & \textbf{0.973}                & \textbf{0.783}                & \textbf{0.972}                & \textbf{0.806}  \\
            \bottomrule
        \end{tabular}
        }
        \caption{}
       \label{tab:diffpretrained_ins}
    \end{subtable}%
    \label{tab:diffpretrained}
    \vspace{-6mm}
\end{table}

\subsection{Improving different pretrained representations}
\label{sec:otherinitializations}
In this section, we show that ItS2CLR is capable of improving representations learned by different pretraining methods: supervised training on ImageNet and two non-contrastive SSL methods, BYOL~\cite{NEURIPS2020_f3ada80d} and DINO~\cite{dino}. 
DINO is by default based on the ViT-S/16 architecture~\cite{dosovitskiy2020image}, whereas the other methods are based on ResNet-18.

Table~\ref{tab:diffpretrained_bag} shows the result of initializing ItS2CLR with pretrained weights obtained from these different models (as well as from SimCLR). The  non-contrastive SSL methods fail to learn better representations than SimCLR. Non-contrastive SSL methods do not use the negative samples, Wang \etal~\cite{wang2021solving} report that this can make the representations under-clustering and result in different object categories overlapping in the representation space. 
The significant improvement of ItS2CLR on top of different pretraining methods demonstrates that the proposed framework is more effective in learning discriminative representations than altering SSL pretraining methods. 

As shown in Table~\ref{tab:diffpretrained_ins}, different initializations achieve varying degrees of pseudo label accuracy, but ItS2CLR improves the performance of all of them. This further demonstrates the robustness of the proposed framework. 

\subsection{Computational complexity}
\label{app:computationtime}
ItS2CLR only requires a small increment in computational time, with respect to existing approaches. For Camelyon16, it takes 600 epochs (approximately 90 hours) to train SimCLR. It only takes 50 extra epochs (approximately 10 hours) to finetune with ItS2CLR, which is only 1/10 of the pretraining time. Updating the pseudo labels is also efficient: it only takes around 10 minutes to update the instance features and training the MIL aggregator. These updates occur every 5 epochs. More training details are provided in Appendix~\ref{subsec:impl}.
\vspace{-1mm}
\section{Related work}
\vspace{-1mm}
\textbf{Self-supervised learning}  Contrastive learning methods have become popular in unsupervised representation learning, achieving state-of-the-art self-supervised learning performance for natural images ~\cite{chen2020simple, he2020momentum, NEURIPS2020_f3ada80d, caron2020unsupervised, zbontar2021barlow, dino}. These methods have also shown promising results in medical imaging \cite{li2021dual, azizi2021big, kaku2021intermediate, zhu2021interpretable,ciga2022self}. Recently, Li \etal~\cite{li2021dual} applied SimCLR~\cite{chen2020simple} to extract instance-level features for WSI MIL tasks and achieved state-of-the-art performance. However, Arora \etal~\cite{arora2019theoretical} point out the potential issue of class collision in contrastive learning, i.e. that some negative pairs may actually have the same class. Prior works on alleviating class collision problem include reweighting the negative and positive terms with class ratio~\cite{chuang2020debiased}, pulling closer additional similar pairs~\cite{dwibedi2021little}, and avoiding pushing apart negatives that are inferred to belong to the same class based on a similarity metric~\cite{zheng2021weakly}. 
In contrast, we propose a framework that leverages information from the bag-level labels to iteratively resolve the class collision problem.

\textbf{Multiple instance learning} A major part of MIL works focuses on improving the MIL aggregator. Traditionally, non-learnable pooling operators such as mean-pooling and max-pooling were commonly used in MIL {~\cite{pinheiro2015image,feng2017deep}}. 
More recent methods parameterize the aggregator using neural networks that employ attention mechanisms~\cite{ilse2018attention,li2021dual,shao2021transmil,transformermil}. 
This research direction is complementary to our proposed approach, which focuses on obtaining better instance representations, and can be combined with different types of aggregators (see Section~\ref{subsec:aggregator}).

\textbf{Self-paced Learning} The core idea of self-paced learning is the ``easy-to-hard'' training scheme, which has been used in semi-supervised learning~\cite{zhang2021flexmatch}, learning with noisy label, unsupervised clustering~\cite{guo2019adaptive}, domain adaptation~\cite{choi2019pseudo,zhang2017curriculum,zhao2020unsupervised,ge2020self}. In this work, we apply self-paced learning to instance representation learning in MIL tasks.

\section{Conclusion}
\label{sec:conclusion}
In this work, we investigate how to improve feature extraction in multiple-instance learning models. We identify a limitation of contrastive self-supervised learning: class collision hinders it from learning discriminative features in class-imbalanced MIL problems. 
To address this, we propose a novel framework that iteratively refines the features with pseudo labels estimated by the aggregator. Our method outperforms the existing state-of-the-art MIL methods on three medical datasets, and can be combined with different aggregators and pretrained feature extractors.

The proposed method does not outperform a cross-entropy-based baseline at very high witness rates, suggesting that it is mostly suitable for low witness rates scenarios (however, it is worth noting that this is the regime more commonly encountered in medical applications such as cancer diagnosis). In addition, there is a performance gap between our method and finetuning using instance-level ground truth, suggesting further room for improvement. 

\noindent{\small{ 
\noindent \textbf{Acknowledgments} The authors gratefully acknowledge NSF grants OAC-2103936 (K.L.), DMS-2009752 (C.F.G., S.L.), NRT-HDR-1922658 (K.L., S.L., Y.S., W.Z.), the NYU Langone Health Predictive Analytics Unit (N.R., W.Z.), the Alzheimer's Association grant AARG-NTF-21-848627 (S.L.), the NIH grant P41EB017183, and the Gordon and Betty Moore Foundation (9683) (K.G., Y.S.).}
}
{\small
\bibliographystyle{ieee_fullname}
\bibliography{egbib}
}
\newpage
\appendix

\begin{center}
    {\Large
\center
\onecolumn 
\textbf{ Appendix for ``Multiple Instance Learning via Iterative Self-Paced
Supervised Contrastive Learning''}}
    \end{center}

\bigskip

The appendix is organized as follows:
\begin{itemize}[leftmargin=*]
    \item In Appendix~\ref{app:experiments}, we include additional descriptions of the datasets (Appendix~\ref{append:dataset}), implementation details (Appendix~\ref{subsec:impl}) and instructions on how to transform the Camelyon16 dataset for the additional experiments (Appendix~\ref{app:subbag}). In Appendix
    ~\ref{app:hyperparameter}, we describe the hyperparameter selection process and report results from an ablation study on the Camelyon16 dataset to evaluate the sensitivity of our approach to the choice of hyperparameters. In Appendix~\ref{Detailsablationmodel}, we provide a detailed description of the ablated versions of ItS2CLR and CE finetuning from Section~\ref{subsec:e2e}. 
    \item In Appendix~\ref{app:results}, we include additional results. In Appendix~\ref{app:learning_curves}, we report pseudo label accuracy  measured by additional metrics between ItS2CLR and the ablated versions. In Appendix~\ref{app:ins_eval}, we show additional results for instance-level performance. In Appendix~\ref{append:e2e}, we report additional comparisons with end-to-end methods. In  Appendix~\ref{app:localization_map}, we provide additional examples of tumor localization maps.
    \item In Appendix~\ref{app:mil_agg}, we provide the formulation and implementation details for the different MIL aggregators used in our study.
\end{itemize}

\section{Experiments}
\label{app:experiments}
\subsection{Dataset}
\label{append:dataset}

\paragraph{Camelyon16} Camelyon16 is a public dataset for detection of metastasis in breast cancer. This dataset consists of 271 training and 129 test whole slide images (WSI). All the images (including both training and test) are  divided into 0.25 million patches at 5× magnification. On average, each slide contains approximately 625 patches 5× magnification respectively. Each WSI is paired with pixel-level annotations indicating the position of tumors (if any are present). We ignore the pixel-level annotations during training and consider only slide-level labels (i.e. the slide is considered positive if it contains any annotated tumor regions). As a result, positive bags contain patches with tumors and patches with healthy tissue.  Negative bags contain only patches with healthy tissue. The ratio between positive and negative 
patches in this dataset is highly imbalanced. Only a small fraction of patches in the positive slides contain tumors (less than 10\%).

\paragraph{TCGA-LUAD} 
TCGA for Lung Adenocarcinoma (LUAD) is a subset of TCGA (The Cancer Genome Atlas), a landmark cancer genomics program. It consists of 800 tumorous frozen whole-slide histopathology images and the corresponding genetic mutation status. Each WSI is paired with a single binary label indicating whether each gene is mutated or wild type. In this experiment, we build MIL models to detect four mutations - EGFR, KRAS, STK11, and TP53, which are sensitizing mutations that can impact treatment options in LUAD~\cite{Coudray197574, Fu813543}. We split the data set randomly into training, validation and test sets so that each patient will appear in only one of the subsets. After splitting the data, 477 images are in the training set, 96 images are in the validation set, and 227 images are in the test set. 

\begin{figure}[t]
    \centering
    \includegraphics[height=0.23\textwidth]{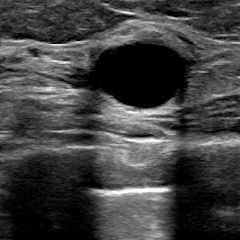}
    \includegraphics[height=0.23\textwidth]{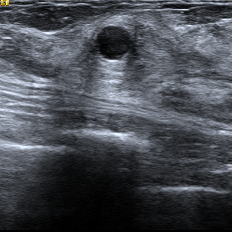}
    \includegraphics[height=0.23\textwidth]{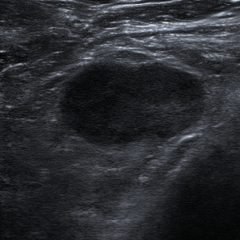}
    \includegraphics[height=0.23\textwidth]{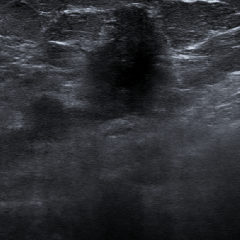}
    \caption{Example breast ultrasound images. The first two images contain a benign lesion. The second and third contain a malignant lesion. In all ultrasound images, the center object of the circular shape corresponds to the lesion of interest. The images are from different exams.}
    \label{fig:example_us}
\end{figure}

\paragraph{Breast Ultrasound dataset} The Breast Ultrasound Dataset includes 28,914 ultrasound exams~\cite{shen2021artificial}. An exam is labeled as cancer-positive if there is a pathology-confirmed malignant finding associated with this exam. In this dataset, 5,593 exams are cancer-positive. On average, each exam contains approximately 18 images. Patients in the dataset were randomly divided into a training set (60$\%$), a validation set (10$\%$), and test set (30$\%$). Each patient was included in only one of the three sets. We show 5 example breast ultrasound images in Figure~\ref{fig:example_us}.

\subsection{Implementation Details}
\label{subsec:impl}

All experiments were conducted on NVIDIA RTX8000 GPUs and NVIDIA V100 GPUs. For all models, we perform model selection during training based on bag-level AUC evaluated on the validation set.

\textbf{Camelyon16} We follow the same preprocessing and pretraining steps as ~\cite{li2021dual}. To preprocess the slides, we cropped the slides into tiles at 5x magnification, filtered out tiles that do not contain enough tissues (average saturation $<$ 30), and resized the images to a resolution of 224 $\times$ 224 pixels. Resizing was performed using the Pillow package~\cite{clark2015pillow} with default settings (nearest neighbor sampling). 

We pretrain the feature extractor, ResNet18~\cite{he2016deep}, with SimCLR~\cite{chen2020simple} for a maximum of 600 epochs, at which we notice that the training loss converges.   Each patch is represented by a 512-dimensional vector. To evaluate the learned representation on the down-stream task, we train a MIL aggregator based on instances representations and evaluate the bag-level prediction every few epochs. We observe that the bag-level AUC on the downstream task in the validation set does not improve when pretraining for a longer period.  We set the batch size to 512 and temperature to 0.5. We use SGD with the learning rate of 0.03, weight decay of 0.0001, and cosine annealing scheduler.

During finetuning the feature extractor with Its2CLR, we finetune the feature extractor for a maximum of 50 epochs. We choose the model that achieves the highest bag-level AUC on the downstream task in the validation set. The batch size is set to 512, and the learning rate is set to $10^{-2}$. At the feature extractor training stage, we apply random data augmentation to each instance, including:
\begin{itemize}
    \item random ($p=0.8$) color jittering:  brightness, contrast, and saturation factors are uniformly sampled from $[0.2,1.8]$, hue factor is uniformly sampled from $[-0.2,0.2]$.,
    \item random grayscale ($p=0.2$),
    \item random Gaussian blur with a kernel size of 0.06 times the size of an image,
    \item random horizontal/vertical flipping with 0.5 probability.
\end{itemize}

When training the DS-MIL aggregator, we follow the settings in~\cite{li2021dual}. We use the Adam optimizer during training. Since each bag may contain a different number of instances, we follow~\cite{li2021dual} and set the batch size to just one bag. We train each model for a maximum of 350 epochs. We use an initial learning rate of $2\times 10^{-4}$, and use the StepLR scheduler to reduce the learning rate by 0.5 every 75 epochs. Details on the hyperparameters used for training the aggregator are in Appendix~\ref{app:mil_agg}.

\textbf{TCGA-LUAD} To preprocess the slides, we cropped them into tiles at 10x magnification, filtered out the background tiles that do not contain enough tissues (when the average saturation is less than 30), and resized the images into the resolution of 224 $\times$ 224 pixels. Resizing was performed using the Pillow package~\cite{clark2015pillow} with nearest neighbor sampling. These tiles were color-normalized with the Vahadane method~\cite{Vahadane}. 

To train the feature extractor, we perform the same process as for Camelyon16. 

We also use DS-MIL~\cite{li2021dual} as the aggregator. When training the aggregator, we resample the ratio of positive and negative bags to keep the class ratio balanced. We train the aggregator for a maximum of 100 epochs using the Adam optimizer with the learning rate set to $2\times 10^{-4}$ and divide the learning rate by 2 every 50 epochs.

\textbf{Breast Ultrasound} We follow the same preprocessing steps as ~\cite{shen2021artificial}. All images were resized to 224 $\times$ 224 pixels using bilinear interpolation. We used ResNet18~\cite{he2016deep} as the feature extractor and pretrained it using SimCLR~\cite{chen2020simple} for 100 epochs, at which we notice that the training loss converges. We adopt the same pretraining and model selection approach as for Camelyon16. We used the Instance Attention-MIL as an aggregator~\cite{ilse2018attention}. Given a bag of images $x_1, ..., x_k$ and a feature extractor $f$, the aggregator first computes instance-level predictions $\hat{y}_i$ for each image $x_i$. It then calculates an attention score $\alpha_i \in [0,1]$ for each image $x_i$ using its feature vectors $f(x_i)$. Lastly, the bag-level prediction is computed as the average instance prediction weighted by the attention score $\hat{y} = \sum_i^k \alpha_{i} \hat{y}_i$. To optimize the aggregator, we trained it using Adam with a learning rate set to $10^{-3}$ for a maximum of 350 epochs. The model is selected according to the best bag-level AUC on the validation set.

\begin{figure}[t]
    \centering
    \includegraphics[height=0.26\textwidth]{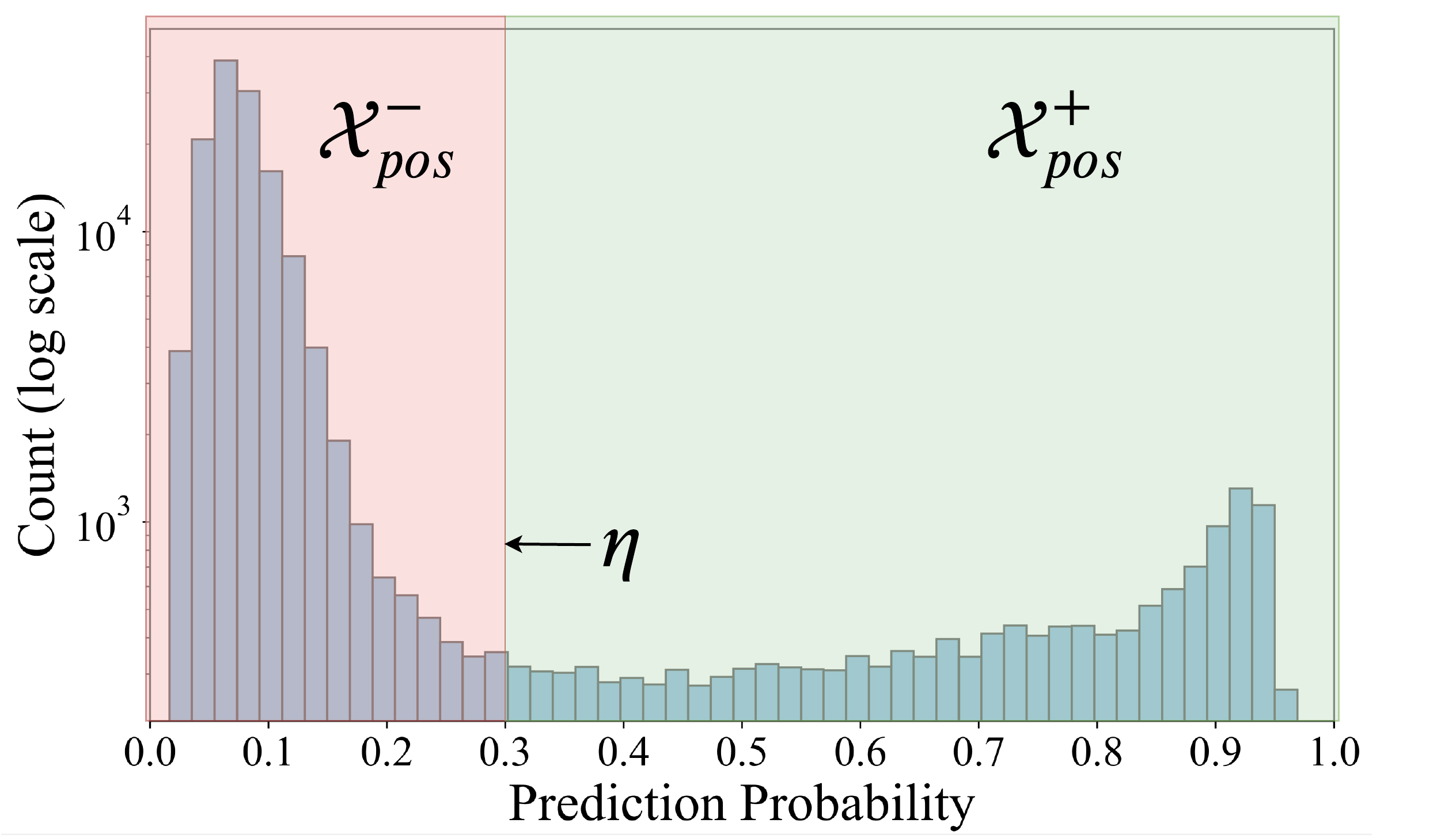}\\
    \vspace{0.8cm}
            \includegraphics[height=0.26\textwidth]{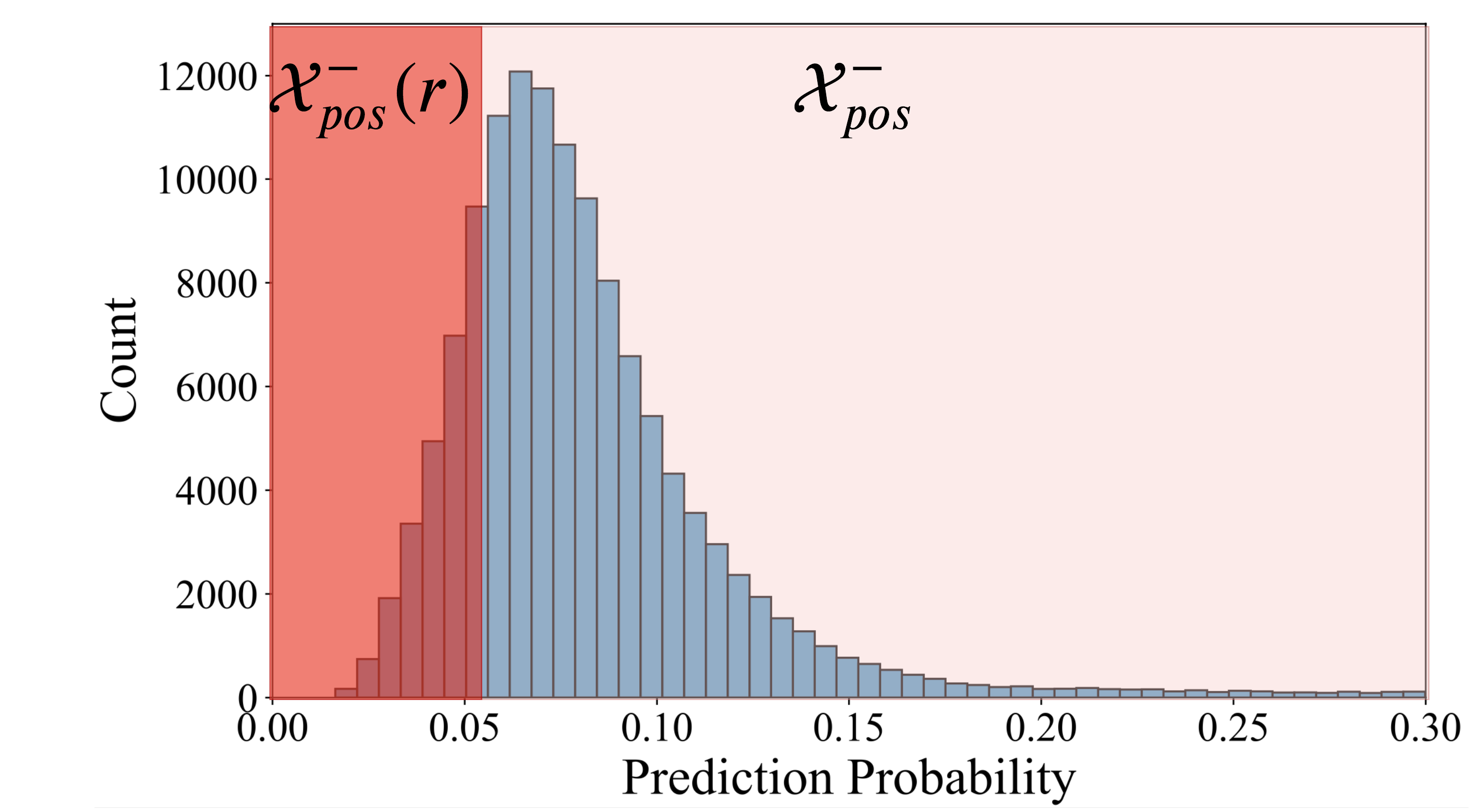}   \includegraphics[height=0.26\textwidth]{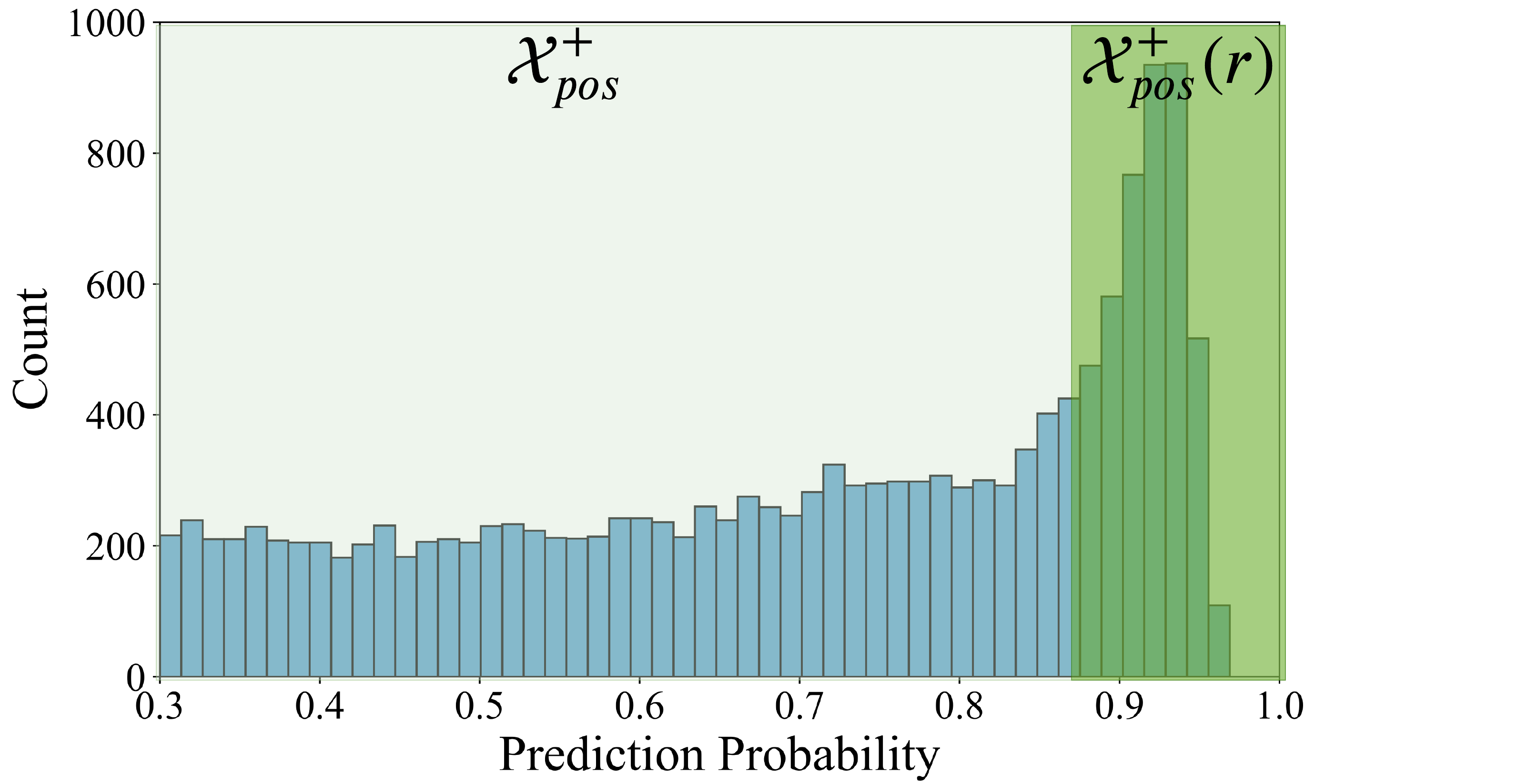} 
    \caption{
    Illustration of our partitioning of the instances from positive bags in Section~\ref{sec:self-paced} based on the predicted probability of the instance classifier in ItS2CLR. \textbf{Top}: $\mathcal{X}^+_{\text{pos}}$ and $\mathcal{X}^-_{\text{pos}}$ are partitioned according to the thresholding parameter $\eta$. \textbf{Bottom}: The distribution of instance scores for instances with negative pseudo labels (left) and negative pseudo labels (right). A threshold $r$ is symmetrically applied on both distributions so that the top $r\%$ instances with the lowest and highest scores are treated as confidently negative or positive, respectively. We use $\mathcal{X}^-_{\text{pos}}(r)$ and $\mathcal{X}^+_{\text{pos}}(r)$ to denote the set of instances that are deemed truly negative and positive respectively.
    During training, as the accuracy of the pseudo labels improves, we can increase $r$ to incorporate more samples in these sets. }
    \label{fig:wholeset}
\end{figure}

\subsection{Hyperparameters for Training the Feature Extractor in ItS2CLR}
\label{app:hyperparameter}

\paragraph{Hyperparameter tuning}
The hyperparameters of the proposed method include: the learning rate $lr \in [1 \times 10^{-5}, 1 \times 10^{-3}]$, the initial threshold used for binarization of the prediction to produce pseudo labels $\eta \in [0.1, 0.9] $, the proportion of sampled positive anchors $p_{+} \in [0.05, 0.5]$, the initial value $r_0 \in [0.01, 0.7]$ and the terminal value $r_T \in [0.2,  0.8]$ of the percentage of selected instances $r$ in the self-paced sampling scheme. For Camelyon16, we obtain the highest bag-level validation AUC using the following hyperparameters: $\eta = 0.3$, $p_{+} = 0.2$, $r_0 = 0.2$ and $r_T = 0.8$. We use the feature extractor trained under these settings in Tables~\ref{tab:mainresult},~\ref{tab:aggregator_results} and \ref{tab:instance_test}. The complete list of hyperparameters in our experiments is reported in Table~\ref{Table:hyperparameter}.

\begin{table}[ht]
    \centering
    \caption{ItS2CLR hyperparameters used in our experiments.}
    \renewcommand{\arraystretch}{1}
    \begin{tabular}{l|c|c|cccc}
    \toprule
    & \multirow{2}{*}{Camelyon16} & Breast & \multicolumn{4}{c}{TCGA-LUAD mutation}\\
    & & Ultrasound &  EGFR  & KRAS  & STK11 & TP53\\
    \midrule
    $\eta$  & 0.3 & 0.3  & 0.3 & 0.5 & 0.3 & 0.5 \\
    $r_0$ & 0.2 & 0.2 & 0.2 & 0.2 & 0.2 & 0.2 \\
    $r_T$ & 0.8 & 0.8 & 0.8 & 0.8 & 0.8 & 0.8 \\
    $p_+$ & 0.2 & 0.2 & 0.5 & 0.2 & 0.2 & 0.2\\
    \bottomrule
    \end{tabular}
    \label{Table:hyperparameter}
\end{table}

\begin{figure}[t]
    \centering
    \includegraphics[width=0.49\textwidth]{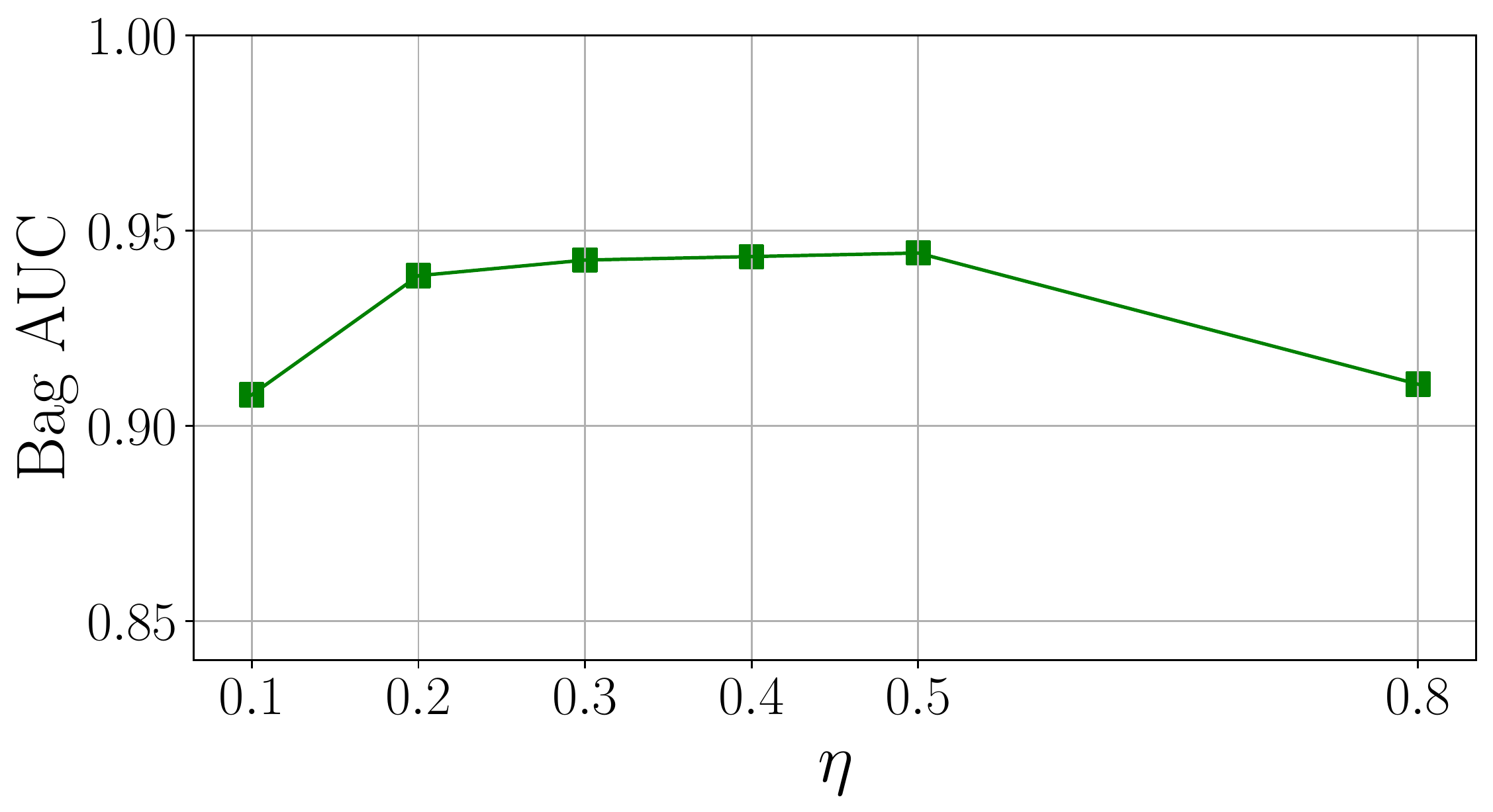}
    \includegraphics[width=0.49\textwidth]{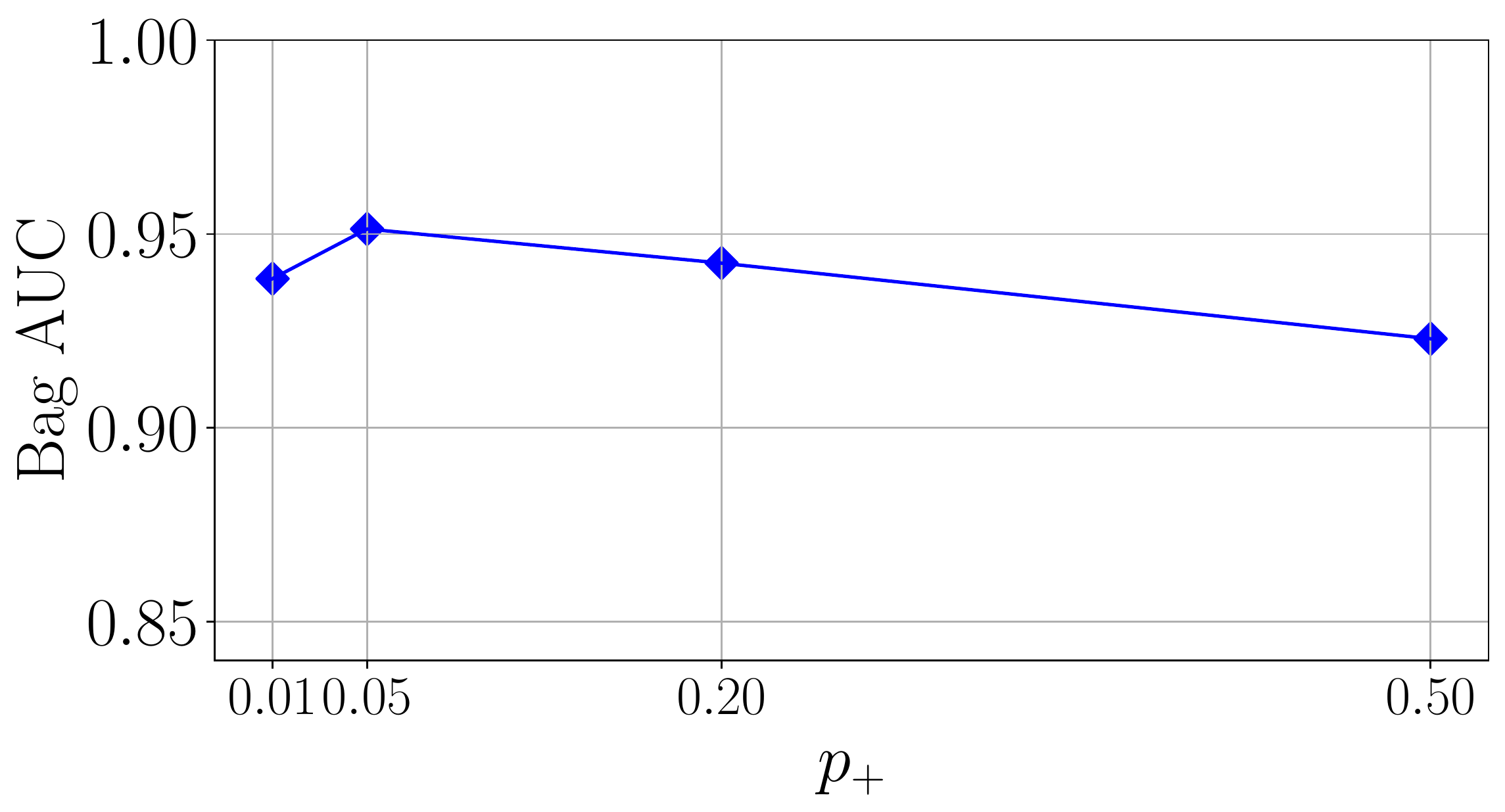}
    \caption{Sensitivity analysis for the threshold $\eta$ and the ratio of positive pseudo labels used as anchor images \smash{$p_{+}$} on the Camelyon16 dataset. }
    \label{fig:ablation_1}
\end{figure}

\begin{figure}[t]
    \centering
    \includegraphics[width=0.49\textwidth]{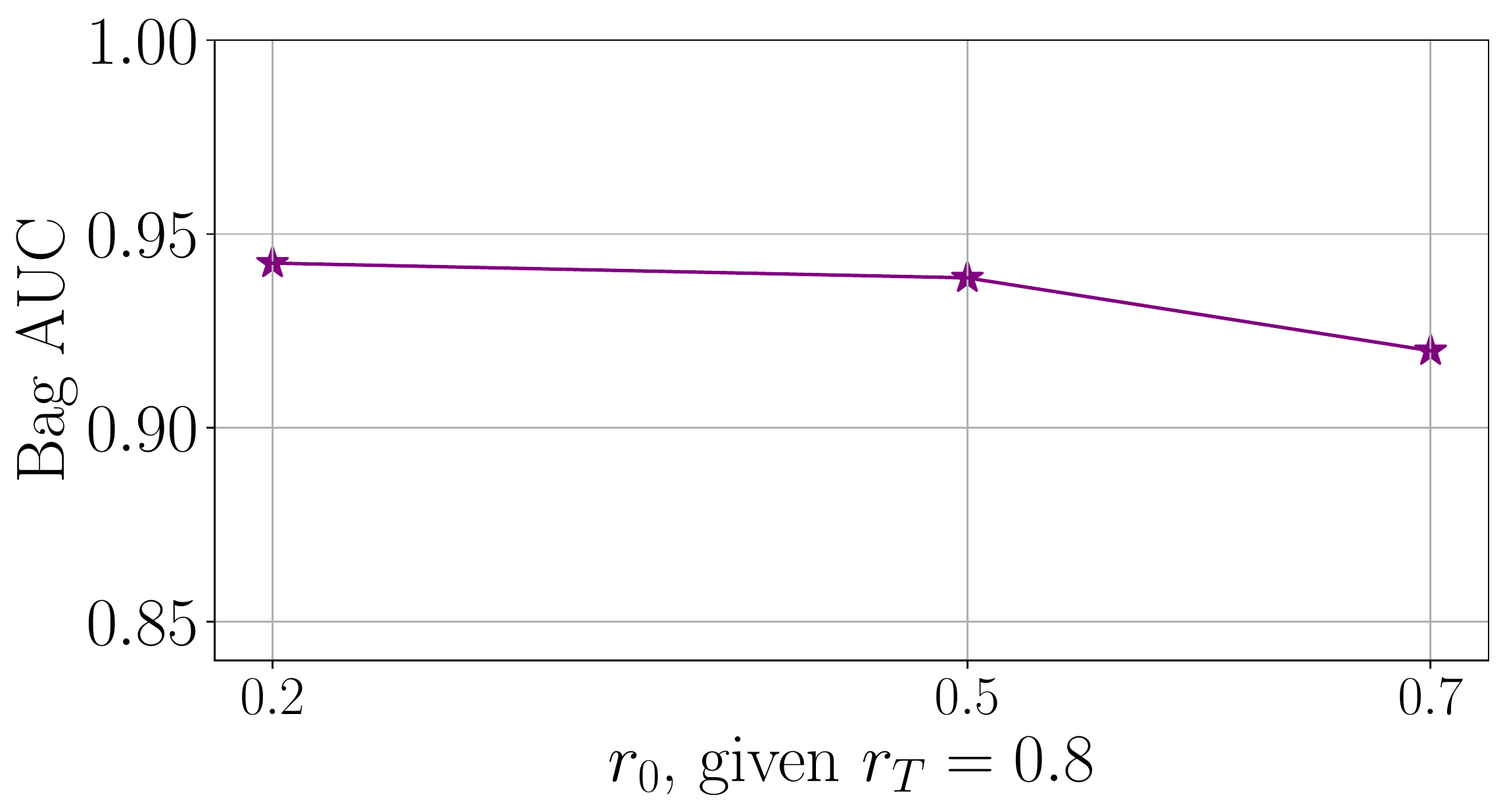}
    \includegraphics[width=0.49\textwidth]{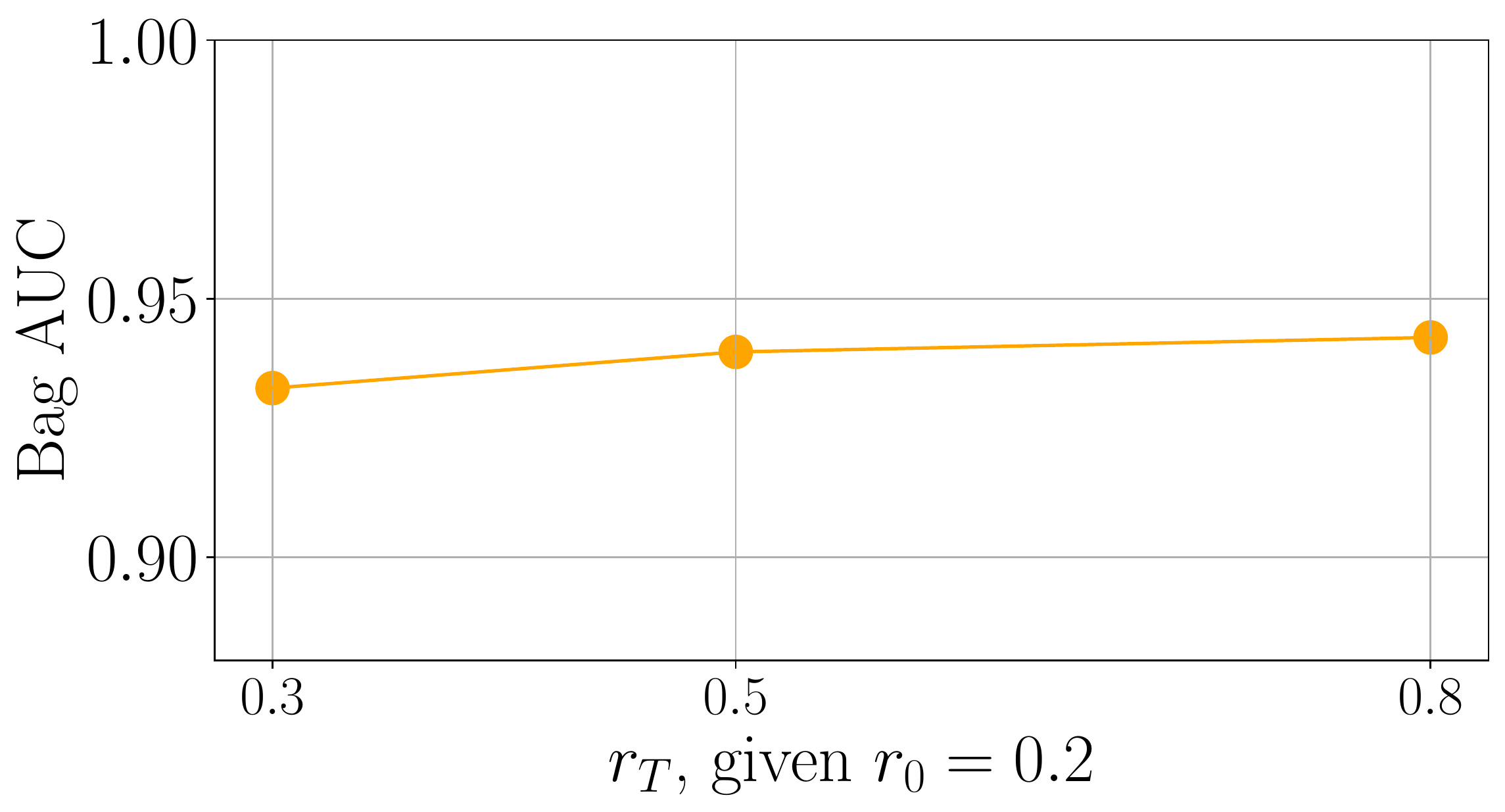}
    \caption{Sensitivity analysis for the hyperparameters $r_0$ and $r_T$ of the proposed self-paced learning scheme on Camelyon16.}
    \label{fig:ablation_2}
\end{figure}

\paragraph{Sensitivity analysis} We conduct a sensitivity analysis for each hyperparameter on Camelyon 16, and observe a robust performance over a range of hyperparameter values. 
\begin{itemize}[leftmargin=*]
    \item \textit{Threshold $\eta$}: 
    The choice of $\eta$ influences the instance-level pseudo labels. As shown in Figure~\ref{fig:wholeset}, the outputs of the instance-level classifier are mostly close to 0 or 1, so the pseudo labels do not dramatically vary for a wide range of $\eta$. 
    We analyze the importance of $\eta$. The left panel of Figure~\ref{fig:ablation_1} shows that ItS2CLR is quite robust to the value of $\eta$, except for extreme values. If $\eta$ is too small (e.g. 0.1), it can introduce a significant number of false positives. If $\eta$ is too large (e.g. 0.8), it can mistakenly exclude some useful positive samples, causing a drop in the performance. 
    In the main paper, since negative instances are more prevalent than positive instances, a threshold of 0.3 (less than 0.5) can increase the recall for the positive instances.  
    \item \textit{Sampling ratio of anchor instance over pseudo labels}: We use $p_{+}$ to denote the percentage of positive anchor instances sampled during the contrastive learning stage. The right panel of Figure~\ref{fig:ablation_1} shows that it is desirable to choose a relatively small $p_{+}$.  Since there are far fewer positive instances than negative instance, keeping the ratio of positive anchors low can avoid repetitively sampling from a limited number of positive instances. Also, since the instance pseudo labels $\X^-_{\text{neg}}$ must be correct by the definition of negative bags, the negative instance pseudo labels are more accurate than the positive ones.
    
    \item \textit{The initial rate $r_0$ and final rate $r_T$ for the self-paced sampling scheduler}: Figure~\ref{fig:ablation_2} shows that ItS2CLR is also generally robust to the values of $r_0$ and $r_T$. However, an extremely large initial rate $r_0$ (high confidence in the pseudo labels) may introduce more samples with incorrect labels during training and hurt the performance. Conversely, extremely small $r_T$ (low confidence in the pseudo labels) may hinder the model from using more data, also hurting performance. 
    \item \textit{Sampling during warm-up:} During the warmup phase, we sample anchor instances from $\mathcal{X}_{\text{neg}}^-$. An alternative choice can be sampling the anchor instance from $\mathcal{X}_{\text{pos}}^+$ and the corresponding set $\mathcal{D}_x$ from $\mathcal{X}_{\text{neg}}^-$. However, our experiments show that the resulting bag-level AUC drops to $90.91$ under this setting, which is significantly lower than $94.25$ by the proposed method. This comparison demonstrates the importance of using clean negative instances as anchor images during warmup. 
\end{itemize}

\subsection{Experiments on Synthetic Versions of Camelyon16} 
\label{app:subbag}

\textbf{Simulation of witness rates (WR)}

Since the ground truth instance-level labels are available for Camelyon16, we can conduct controlled experiments on synthetic versions of the dataset. We manipulate the prevalence of positive instances in the bag (the \emph{witness rate}) and study its impact on the performance of the proposed approach and the baselines, as reported in Section~\ref{sec:motivation}.  
To increase the witness rate, we randomly drop the negative instances at a fixed ratio in each bag; to reduce the witness rate, we randomly drop the positive instances at a fixed ratio in each bag. 
The percentage of retained instances and the resulting witness rates are reported in Table~\ref{tab:WR_exp}.

\textbf{Downsampled version of Camelyon16 for end-to-end training}

In order to enable end-to-end training, we downsample each bag in Camelyon16 to around 500 instances so that it fits in the memory of a GPU. To achieve this, we divide large bags which have more than 500 instances into smaller bags.

For negative bags, we randomly partition the instances within the original bag into same-sized sub-bags with around 500 instances. 

For positive bags, we randomly partition the positive instances within the original bag into the desired number of sub-bags. We adjust the number of sub-bags so that it cannot be less than the number of positive instances. We then combine the positive instances and the negative instances to form sub-bags. This ensures that the bag-level label is correct and the witness rate for each positive sub-bag remains similar to the original bag.

\subsection{Description of the ablation study}
\label{Detailsablationmodel}

\textbf{Details for CE finetuning with/without iterative updating}
\begin{itemize}[leftmargin=*]
    \item \textit{CE without iterative updating:}  we use the same initial pseudo labels and pretrained representations as our ItS2CLR framework. Concretely, we label all the instances in negative bags as negative. We label the instances in positive bags using the instance prediction obtained from the aggregator. When finetuning  the aggregator, the pseudo labels are kept fixed. 
    \item\textit{ CE + iterative updating:}  based on CE with iterative updating, the pseudo labels are updated every few epochs, which is in turn used to guide the finetuning of the feature extractors. 
\end{itemize}

\textbf{Details for ItS2CLR with/without SPL}

\begin{itemize}[leftmargin=*]
\item \textit{ItS2CLR without iterative updating:} we keep everything the same as the full Its2CLR procedure (including the SPL strategy), but we do not apply steps 7, 8 and 9 in Algorithm~\ref{algo:main}. As a result, the pseudo labels are fixed to the initial set of pseudo labels. 
\item \textit{ItS2CLR without SPL:} we keep everything the same as the full Its2CLR procedure (including iterative updating), but modify step 10 in Algorithm~\ref{algo:main}. We do not utilize the pseudo label to train the model in a self-paced learning way as in Section~\ref{sec:self-paced}. We utilize all the pseudo-labeled data from the beginning of the finetuning.  
\end{itemize}

\section{Additional Results}
\label{app:results}

\subsection{Learning Curves}
\label{app:learning_curves}

\begin{figure}[t]
    \centering
    \includegraphics[width=0.7\textwidth]{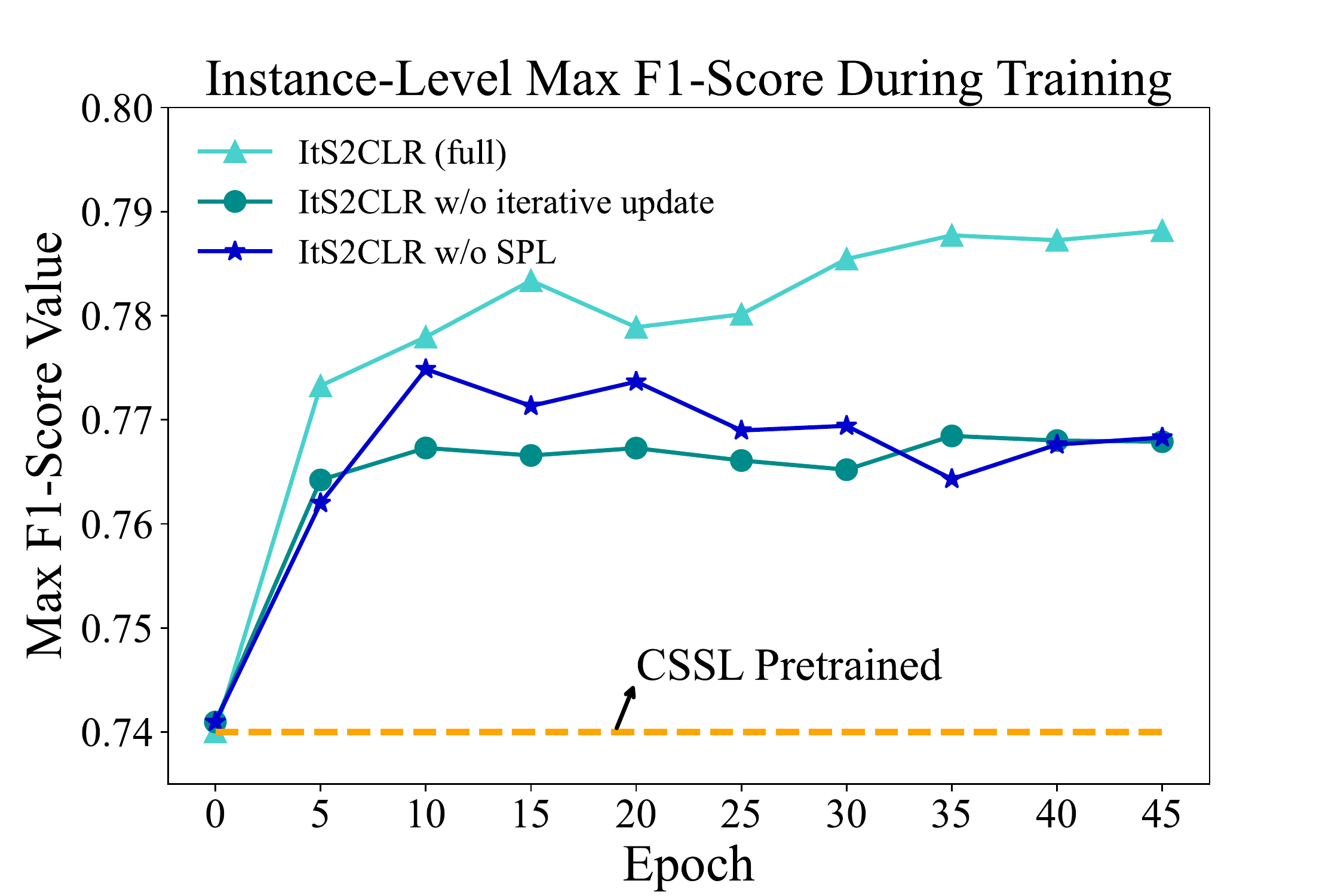}
    \caption{Comparison of max F1 score on instance pseudo labels. ItS2CLR updates the features iteratively based on a subset of the pseudo labels that are selected according to the self-paced learning (SPL) strategy. On Camelyon16, this gradually improves the accuracy of the pseudo labels in terms of instance-level max F1 score. Both the iterative updates and SPL are important to achieve this.}
    \label{fig:pseudolabelf1}
\end{figure}

\textit{F1-Score plot corresponding to Figure~\ref{fig:training_process}:} In Figure~\ref{fig:pseudolabelf1}, we show the max F1 score curve corresponding to the right side of Figure~\ref{fig:training_process}. This plot confirms the importance of self-paced learning and iterative updating in ItS2CLR.

\begin{figure}[ht]
    \centering
    \includegraphics[width=0.7\textwidth]{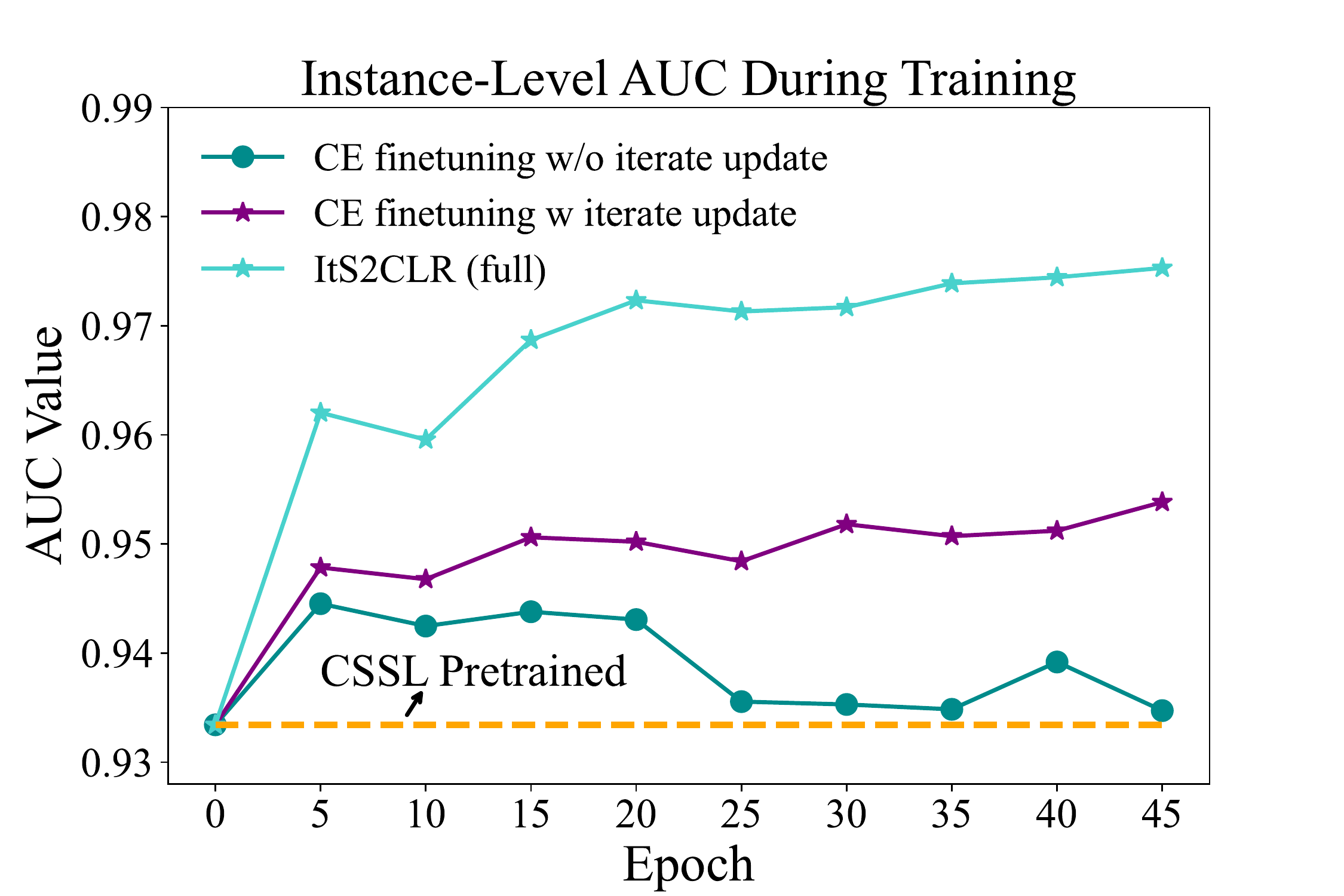}
    \caption{Comparison of instance-level AUC during training between ItS2CLR and an alternative approach that finetunes the feature extractor using cross-entropy (CE) loss on the Camelyon16 dataset. Iterative updating improves performance for CE finetuning, but ItS2CLR produces more accurate pseudo labels.
    }
    \label{fig:pseudolabelCE}
\end{figure}

\textit{Instance-level AUC during training comparison with cross-entropy finetuning:} Figure~\ref{fig:pseudolabelCE} compares ItS2CLR with an alternative approach that finetunes the feature extractor using cross-entropy (CE) loss on the Camelyon16 dataset. Without iterative updating, CE finetuning rapidly overfits to the incorrect labels. Iterative updating prevents this to some extent but does not match the performance of ItS2CLR, which produces increasingly accurate pseudo labels as the iterations proceed.

\begin{figure}[ht]
\vspace{-3mm}
\centering
\includegraphics[width=0.8\linewidth]{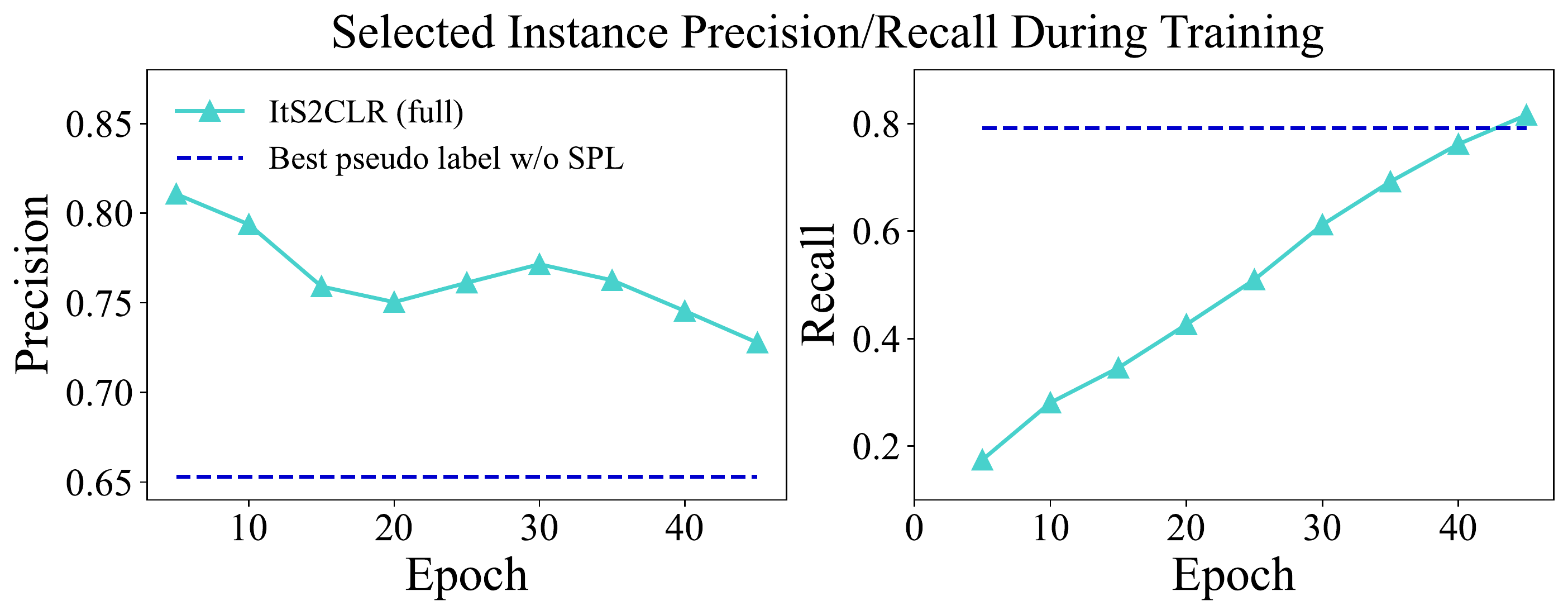}
\caption{The precision and recall of pseudo labels from the selected instances during fine-tuning. The proposed ItS2CLR approach achieves high precision in generating pseudo labels from selected instances during fine-tuning through self-paced sampling. Since instance-level AUC improves during training (as shown in Fig~\ref{fig:training_process}), gradually including more instance candidates leads to higher recall while maintaining significant precision.
}
\label{fig:pseudolabelpr}
\end{figure}

\subsection{Instance-level Evaluation}
\label{app:ins_eval}

In order to evaluate instance-level performance, we report values of classification metrics including AUC, F1-score, AUPRC and Dice score for localization.

The Dice score is defined as follows:
\begin{equation}
    \text{Dice} = \frac{2\sum_i y_i p_i}{\sum_i y_i + \sum_i p_i}, 
\end{equation}
where $y_i$ and $p_i$ are the ground truth and predicted probability for the $i$th sample. 
The predicted probability is computed from the output of the MIL model $s_i$ via linear scaling:
\begin{equation}
p_i = \sigma\left( as_i + b\right),
\end{equation}
where $a \in [-5, 5]$ and $b \in [0.1, 10]$ are chosen to maximize the Dice score on the validation set. 
\begin{table}[t]
    \caption{Comparison of instance-level performance for the models in Table~\ref{tab:aggregator_results}, using a max pooling aggregator.}
    \centering
    \renewcommand{\arraystretch}{1.2}
    \begin{tabular}{lcc|cc|cccc}
    \toprule
     $(\times 10^{-2})$& SimCLR & \multirow{2}{2cm}{\centering{Ground-truth finetuning}} & \multicolumn{2}{c|}{CE finetuning} & \multicolumn{4}{c}{ItS2CLR} \\
    & (CSSL) && \textit{w/o iter.} & \textit{iter.} & \textit{w/o both} & \textit{w/o iter.}  & \textit{w/o SPL}   & \textit{Full}  \\
    \midrule
    AUC &91.53 & 97.58 & 93.17 & 94.48 & 92.69 & 94.55 & 94.43 & \textbf{96.25} \\
    F1-score      &78.45 & 88.24 & 85.26 & 85.83 & \textbf{86.77} & 86.05 & 87.52 & 86.75 \\
    AUPRC  & 79.94 & 85.50 & 85.79 & 86.73 & 85.50 &84.54 & 86.80 & \textbf{89.99} \\
    Dice & 31.21 & 63.01 & 43.90 & 44.76 & 46.57 & 55.30 & 52.55 &  \textbf{57.82}\\
\bottomrule
\end{tabular}

    \label{tab:instance_test2}
\end{table}

\begin{table}[t]
    \caption{Comparison of instance-level performance for the models in Table~\ref{tab:aggregator_results}, using a linear classifier trained on the frozen features produced by each model. In addition, we produce bag-level predictions using the maximum output of the linear classifier for each bag.}
    \centering
    \renewcommand{\arraystretch}{1.2}
    \begin{tabular}{lcc|cc|cccc}
    \toprule
     $(\times 10^{-2})$& SimCLR & \multirow{2}{2cm}{\centering{Ground-truth finetuning}} & \multicolumn{2}{c|}{CE finetuning} & \multicolumn{4}{c}{ItS2CLR} \\
    Instance-level & (CSSL) && \textit{w/o iter.} & \textit{iter.} & \textit{w/o both} & \textit{w/o iter.}  & \textit{w/o SPL}   & \textit{Full}  \\
    \midrule
    AUC &96.13 & 97.56 & 96.94 & 96.88 & 96.64 & 97.25 & 96.92 & \textbf{97.27}\\
    F1-score      &85.29 & 87.69 & 87.34 & 86.94 & 86.00 & 87.07 & 87.6 & \textbf{87.92}\\
    AUPRC  & 82.65 & 85.94 & 79.96 & 78.02 & 78.17 & \textbf{84.56} & 77.90 & 82.09 \\
    Dice  & 49.56 & 61.39 & 55.40 & 54.66 & 51.85 & 54.87 & 55.11 & \textbf{60.13}\\
    \hhline{=========}
    \multicolumn{2}{l}{Bag-level (max-pooling)} &  & & && & & \\
    \midrule
    AUC & 86.25 & 97.37 & 87.53 & 90.54 & 89.97 & 93.09 & 92.81 & \textbf{97.47} \\
\bottomrule
\end{tabular}
    \label{tab:instance_test3}
\end{table}

\begin{figure}[t]
    \centering
    \includegraphics[width=0.6\textwidth]{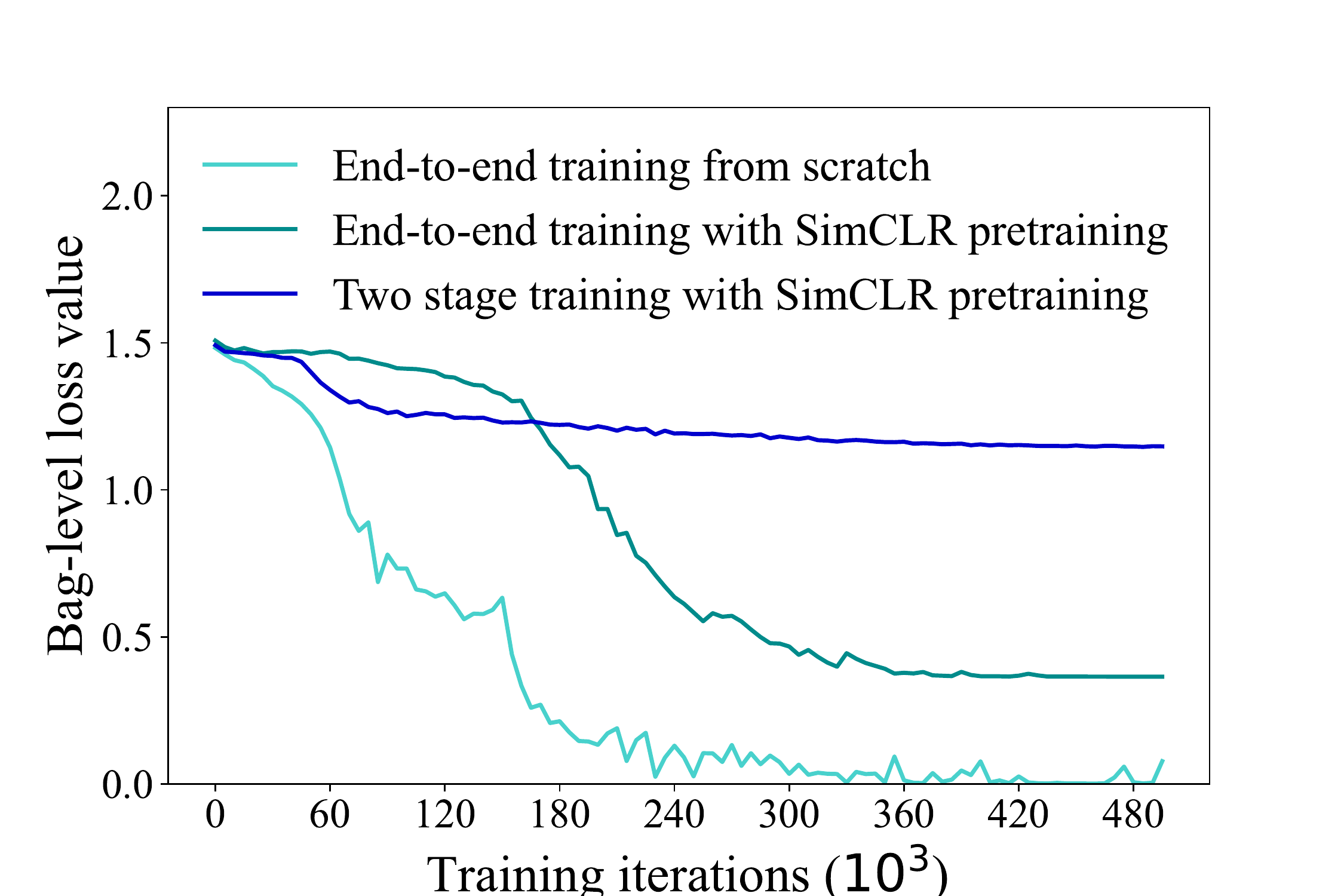}  
    \caption{Comparison between end-to-end training and two-stage training on the downsampled version of the Camelyon16 dataset. End-to-end models overfit rapidly. Note that the unit of the training iterations here is 1k.}
    \label{fig:e2etraining}
\end{figure}

\textbf{Max pooling aggregator}: In Table~\ref{tab:instance_test}, we show that our model achieves better weakly supervised localization performance compared to other methods when DS-MIL is used as the aggregator. In Table~\ref{tab:instance_test2}, we show that the same conclusion holds for an aggregator based on max-pooling.

\textbf{Linear evaluation}: In Table~\ref{tab:instance_test3}, we report results obtained by training a logistic regression model using the features obtained from the same approaches in Table~\ref{tab:instance_test}, following a standard linear evaluation pipeline in representation learning~\cite{chen2020simple}. ItS2CLR again achieves the best instance-level performance. We also produce bag-level predictions using the maximum output of the linear classifier for each bag, which again showcases that better instance-level performance results in superior bag-level classification.

\subsection{Comparison with End-to-end Training}
\label{append:e2e}

In this section, we provide additional results to complement  Table~\ref{tab:e2e_comp}, where ItS2CLR is compared to end-to-end models. The end-to-end training is conducted with the same aggregators for each dataset as described in Section~\ref{sec:experiments} and Appendix~\ref{subsec:impl}.

\begin{table}[t]
    \centering
    \caption{Results on the downsampled version of the Camelyon16 dataset. }
\renewcommand{\arraystretch}{1.2}
\begin{tabular}{>{\centering\arraybackslash}m{0.2\linewidth} | >{\centering\arraybackslash}m{0.15\linewidth} >{\centering\arraybackslash}m{0.15\linewidth} |
>{\centering\arraybackslash}m{0.15\linewidth} >{\centering\arraybackslash}m{0.15\linewidth} }
\toprule & End-to-end (scratch) & End-to-end (SimCLR) & SimCLR + DS-MIL & ItS2CLR \\
\hline Bag AUC & 64.52 & 66.71  & 80.96 & \textbf{88.65} \\
Instance AUC & 78.32& 81.29& 93.94&\textbf{95.58}\\
Instance F-score & 51.02 & 55.71&  85.93&\textbf{87.01}\\
  \bottomrule
\end{tabular}
    \label{tab:e2eappen}
\end{table}

\begin{table}[t]
    \caption{Results on the Breast Ultrasound dataset.}
    \centering
    \renewcommand{\arraystretch}{1.2}
    \begin{tabular}{l|ccc}
    \toprule
         &   SimCLR + Aggregator & End-to-end MIL & ItS2CLR\\
    \midrule
    Bag AUC     &  80.79 & 91.26 & \textbf{93.93} \\
    Bag AUPRC     &  34.63 & 58.73 & \textbf{70.30} \\
    \hline
    Instance AUC    &  62.83 & 82.11 & \textbf{88.63} \\
    Instance AUPRC    &  10.58 & 31.31 & \textbf{43.71} \\
    \bottomrule
    \end{tabular}
    \label{tab:ultrasound}
\end{table}

\textbf{Camelyon16}
Figure~\ref{fig:e2etraining} shows that an end-to-end model trained on the downsampled version of Camelyon16 described in Section~\ref{app:subbag} rapidly overfits when trained from scratch and from SimCLR-pretrained weights. The two-stage model, on the other hand, is less prone to overfitting. Table~\ref{tab:e2eappen} shows that the two-stage learning pipeline outperforms end-to-end training, and is in turn outperformed by ItS2CLR. 

\textbf{Breast Ultrasound dataset}
Table~\ref{tab:ultrasound} shows that end-to-end training outperforms the SimCLR+Aggregator baseline  for the breast-ultrasound dataset, but is outperformed by ItS2CLR.

\subsection{Tumor Localization Maps}
\label{app:localization_map}

\begin{figure}[t]
    \centering
    \includegraphics[width=\textwidth]{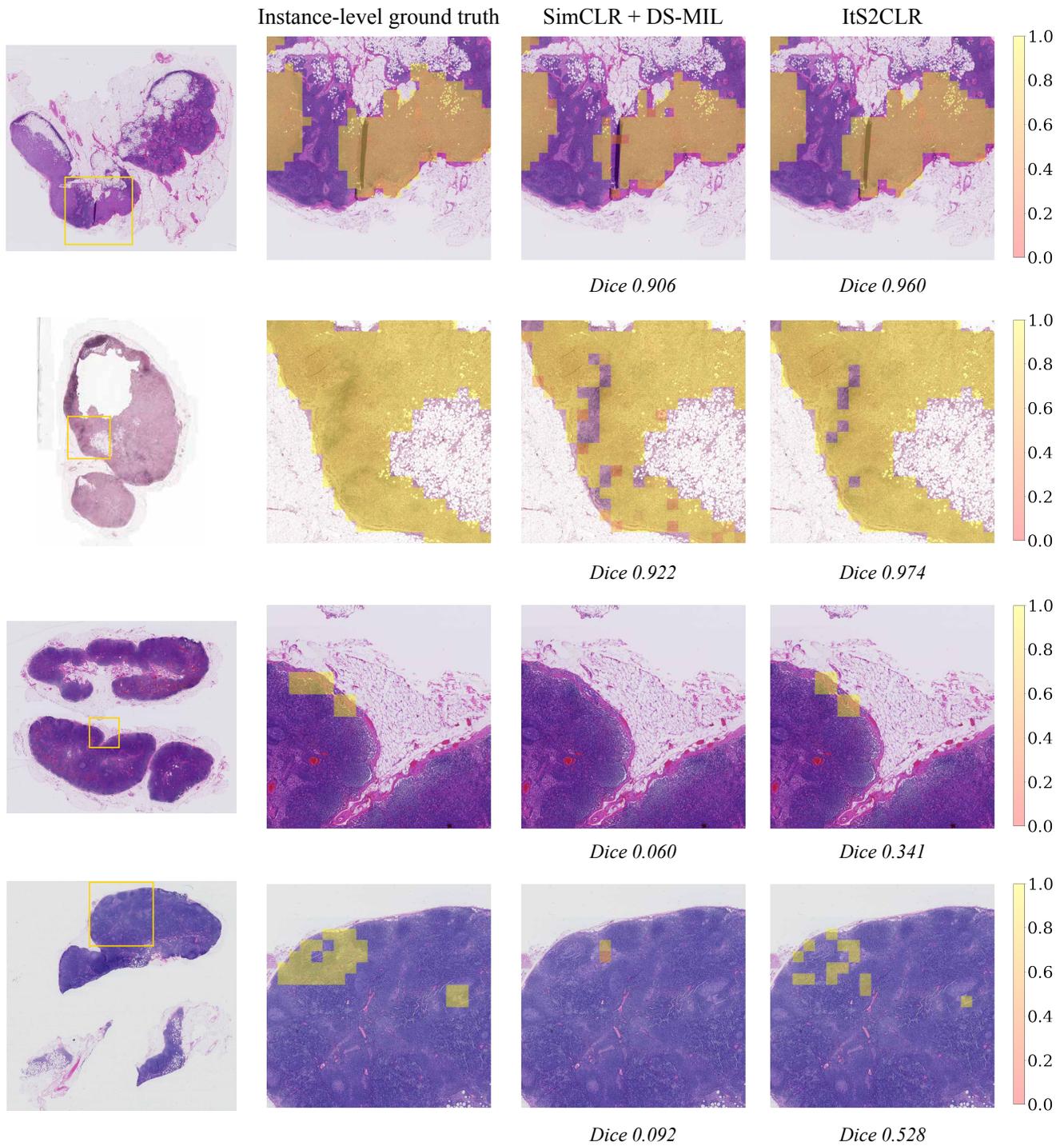}
    \caption{Additional tumor localization maps for histopathology slides from the Camelyon16 test set. Instance-level predictions are generated by the instance-level classifier of the DS-MIL trained on extracted instance-level features.}
    \label{fig:more_loc_maps}
\end{figure}

 Figure~\ref{fig:more_loc_maps} provides additional tumor localization maps. 

\section{MIL Aggregators}
\label{app:mil_agg}

\subsection{Formulation of MIL Aggregators}
In this section, we describe the different MIL aggregators benchmarked in Section~\ref{subsec:aggregator} and Table~\ref{tab:aggregator_results}.

Let $\mathcal{B}$ denote a collection of sets of feature vectors in $\mathbb{R}^d$. The bags of extracted features in the dataset are denoted by $\{H_b\}_{b=1}^{B} \subset \mathcal{B}$. 
An aggregator is defined as a function $g: \mathcal{B} \rightarrow [0, 1]$ mapping bags of extracted features to a score in $[0, 1]$. 

There exist two main approaches in MIL:
\begin{enumerate}
    \item  \textit{The instance-level approach}: using a logistic classifier on each instance, then aggregating instance predictions over a bag (e.g. max-pooling, top k-pooling).
    \item \textit{The embedding-level approach}: aggregating the instance embeddings, then obtaining a bag-level prediction via a bag-level classifier (e.g. attention-based aggregator, Transformer).
\end{enumerate}

We denote the embeddings of the instances within a bag by $H = {\{h_k\}_{k=1}^{K}}$, where $K$ is the number of instances.

\textbf{Max-pooling} obtains bag-level predictions by taking the maximum of the instance-level predictions produced by a logistic instance classifier $\phi$, that is

\begin{equation}
    g_\phi(H) = \max_{k=1, \cdots, K} \left \{ \phi(h_k)\right \}.
\end{equation}

\textbf{Top-k pooling}~\cite{shen2021interpretable} produces bag-level prediction using the mean of the top-$M$ ranked instance-level predictions produced by a logistic instance classifier $\phi$, where $M$ is a hyperparameter. 

Let $\text{top}M \left(\phi, H\right)$ denote the indices of the elements in $H$ for which $\phi$ produces the highest $M$ scores,
\begin{equation}
    g_\phi(H) = \frac{1}{M}\sum_{k\in \text{top}M \left(\phi, H\right)} \phi(h_k).
\end{equation}

\textbf{Attention-based MIL} \cite{ilse2018attention} aggregates instance embeddings using a sum weighted by attention weights. Then the bag-level estimation is computed from the aggregated embeddings by a logistic bag-level classifier $\varphi$:
\begin{equation}
    g_\varphi(H) = \varphi\left( \sum_{k=1}^K a_k h_k \right),
\end{equation}
where $a_k$ is the attention weight on instance $k$:
\begin{equation}
    a_k = \frac{\exp\left(w^T \tanh(Vh_k^T)\right)}{\sum_{j=1}^K\exp\left(w^T \tanh(Vh_j^T)\right)},
\end{equation}
where \smash{$w \in \mathbb{R}^{l \times 1}$} and \smash{$V \in \mathbb{R}^{l \times d}$} are learnable parameters and $l$ is the dimension of the hidden layer.

\textbf{DS-MIL} combines instance-level and embedding-level aggregation, we refer to DS-MIL~\cite{li2021dual} for more details on this approach.

\textbf{Transformer}~\cite{transformermil} proposed an aggregation that uses an $L$-layer Transformer to process the set of instance features $H$. The initial set $H^{(0)}$ is set equal to $H$. Then it goes through the Transformer as follows:
\begin{equation}
    \begin{aligned}
    & H'^{(l)} = \text{MSA}\left(H^{(l - 1)} \right) + H^{(l-1))} \\
    & H^{(l)} = \text{MLP}\left(H'^{(l - 1)} \right) + H'^{(l-1)} \\
    \end{aligned}
\end{equation}
for $l = 1, \cdots, L$, where $\text{MSA}$ is multiple-head self-attention, $\text{MLP}$ is a multi-layer perceptron network. Then the processed vectors $ H^{(l)}$ are fed to \textbf{Attention-based MIL}  \cite{ilse2018attention} to obtain the bag-level predictions

\begin{equation}
    g_\varphi(H^l) = \varphi\left( \sum_{k=1}^K a_k h_k^l \right).
\end{equation}

Here $a_k$ is the attention weight on instance $k$:

\begin{equation}
    a_k = \frac{\exp\left(w^T \tanh(V(h_k^l)^T)\right)}{\sum_{j=1}^K\exp\left(w^T \tanh(V(h_j^l)^T)\right)},
\end{equation}

where \smash{$w \in \mathbb{R}^{p \times 1}$} and \smash{$V \in \mathbb{R}^{p \times d}$} are learnable parameters and $p$ is the dimension of the hidden layer.

\subsection{Implementation Details}

\textbf{Top-k pooling} We select the ratio in Top-k pooling from the set $\{ 0.1\%, 1\%, 3\%, 10\%, 20\% \}$.

\textbf{DS-MIL} The weight between the two cross-entropy loss functions in DS-MIL is selected from the interval $[0.1, 5]$ based on the best validation performance.

\textbf{Attention-based MIL}. The hidden dimension of the attention module to compute the attention weights is set equal to the dimension of the input feature vector (512).

\textbf{Transformer}. We add light-weighted two-layer Transformer blocks to process instance features. We did not observe improvement in  performance with additional blocks.

\end{document}